\documentclass{article}
\usepackage{xcolor}         

\usepackage{newunicodechar}

\newunicodechar{−}{-}

 \usepackage[preprint]{neurips_2026}

\usepackage[utf8]{inputenc} 
\usepackage[T1]{fontenc}    
\usepackage{hyperref}       
\usepackage{url}            
\usepackage{booktabs}       
\usepackage{amsfonts}       
\usepackage{nicefrac}       
\usepackage{microtype}      
\usepackage{xcolor}         

\usepackage{multirow} 
\usepackage{pifont}

\usepackage{multirow}       
\usepackage{nicefrac}       
\usepackage{microtype}      
\usepackage{graphicx}
\usepackage{caption}
\usepackage{xspace}
\usepackage{amsmath}        
\usepackage{algpseudocode}
\usepackage{algorithm}
\usepackage{makecell}
\usepackage{amssymb} 
\usepackage{float}
\usepackage{subcaption}
\usepackage{enumitem} 
\usepackage{listings}
\usepackage{newunicodechar}
\usepackage{placeins}
\title{SceneBind: Binding What and Where Across Vision, Audio and Language}

\author{
  \vspace{-25pt}\\
  \textbf{Mingfei Chen$^{1}$,\quad Zijun Cui$^{1,2}$\thanks{Led the data curation work while at the University of Washington},\quad Ruoke Zhang$^{1}$,\quad Hyeonggon Ryu$^{3}$,\quad Eli Shlizerman$^{1}$\thanks{Corresponding author: \href{mailto:shlizee@uw.edu}{shlizee@uw.edu}}}
  \vspace{3pt} \\
  $^{1}$University of Washington 
  \quad $^{2}$University of Texas at Dallas 
  \\ $^{3}$Hankuk University of Foreign Studies
  \vspace{8pt} \\
  \vspace{-4pt}
}

\begin{document}

\maketitle

\begin{figure}[htb]
    \centering
    \includegraphics[width=0.95\linewidth]{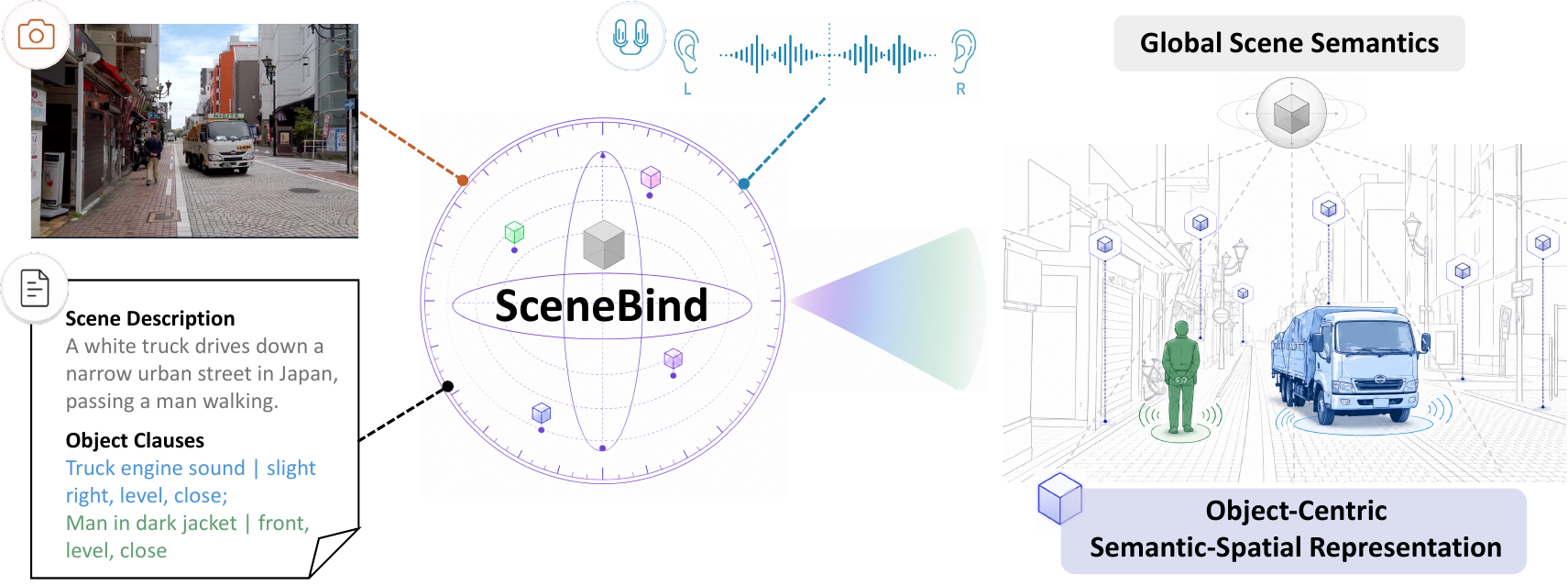}
    \vspace{-1mm}
    \caption{SceneBind models scenes by jointly capturing semantics (\emph{what}) and 3D spatial attributes (\emph{where}) across vision, audio and language. SceneBind maps each modality into a shared global embedding with object-centric slots, enabling semantic–spatial representation of multimodal scenes.
}
    \label{fig:teaser}
\end{figure}

\begin{abstract}
We present \textbf{SceneBind}, an omni-modal representation of realistic scenes with joint semantic and 3D spatial understanding across vision, audio and language. Existing omni-modal encoders excel at instance-level semantics (i.e., \emph{what} is present), but often lack explicit spatial structure (i.e., \emph{where} it is). 
SceneBind addresses this gap by representing each scene as a semantic-spatial entity, combining a global semantic embedding with object-centric semantic-spatial slots. This representation explicitly captures object-level semantics, spatial attributes, and uncertainty.
We further propose \textbf{SceneBind Matching}, a semantic–spatial matching scheme that integrates global scene similarity with object alignment, supporting cross-modal scene retrieval and object grounding.
To train and evaluate SceneBind, we curate a novel real-world binaural audio-visual dataset with structured semantic and spatial annotations, and propose a training protocol for aligning semantic and spatial signals across modalities.
SceneBind is compatible with large-scale pretrained semantic encoders, adds lightweight spatial modeling with only a few additional tokens. It achieves state-of-the-art scene and spatial retrieval while enabling strong zero-shot transfer to downstream tasks such as audio-visual localization.

\end{abstract}

\section{Introduction}
Understanding real world environments emphasizes semantic and spatial perception which captures both \emph{what} is present and \emph{where} it is in 3D space. This capability is essential for multimodal spatial intelligence~\cite{yang2024think,chen2025savvy}, enabling spatial reasoning and interaction across vision, audio and language, and supporting applications in embodied AI, robotics, and spatially grounded multimodal world models.

Recent advances in large-scale multimodal learning have yielded powerful representations~\cite{radford2021learning,Zhai2023SigmoidLF,Tschannen2025SigLIP2M,laion-clap,m2d-clap,girdhar2023imagebind} that align semantic concepts of a scene across modalities such as images and audio. While highly effective for semantic understanding, these models provide limited explicit modeling of spatial structure. In this work, we aim to learn a representation that unifies semantics and spatial structure for real-world omni-modal scenes where each modality is encoded into a shared semantic–spatial space.

Achieving this goal presents several challenges. First, semantic embeddings typically encode global scene content, whereas spatial information is inherently object-centric. In complex scenes, a single embedding is often insufficient to represent multiple objects with distinct spatial configurations. Second, cross-modal spatial learning is data-limited. Real-world datasets have advanced vision-audio-language alignment, but explicit spatial annotations remain limited. Learning to associate a visual object or textual description with a sound source at a particular direction and distance requires spatial audio recordings with aligned source and spatial annotations~\cite{chen2025savvy}. Synthetic acoustic simulators~\cite{chen2022soundspaces} offer useful partial solutions, but still face realism gaps and misalignment with real-world visual data.

To address these challenges, we propose \textbf{SceneBind}, a framework for learning semantic–spatial representations of omni-modal scenes. As shown in Figure~\ref{fig:teaser}, SceneBind represents a 3D scene through a global semantic embedding and object-centric semantic–spatial representations, capturing both holistic context and multi-object spatial structure with compact tokens. To support this formulation, we introduce a data curation pipeline based on binaural videos, constructing a novel omni-modal dataset and benchmark with structured spatial annotations for real-world scenes.

Building on SceneBind representation and dataset, we develop a training strategy for learning object-centric representations with varying numbers of objects in a scene. We use bipartite matching to assign predicted object slots to ground-truth semantic-spatial object descriptions, and combine global semantic alignment, object grounding, and object-centric contrastive objectives to learn discriminative, spatially grounded representations. We further introduce \textbf{SceneBind Matching} for cross-modal scene comparison, enabling retrieval, grounding, and zero-shot tasks such as audio-visual localization.

Our contributions are: (1) We introduce \textbf{SceneBind} for learning omni-modal scene representations that capture semantics (\emph{what} is present) and spatial location (\emph{where} it is) through global and object-centric joint representations.
(2) We curate a novel real-world binaural audio-visual dataset with structured spatial annotations to support training and benchmarking semantic-spatial scene representations.
(3) We propose \textbf{SceneBind Matching}, combining global and object-level alignment for cross-scene comparison, retrieval, and grounding.
(4) SceneBind integrates with pretrained encoders with minimal overhead, achieving state-of-the-art performance and zero-shot generalization.

\section{Related Works}
\subsection{Multimodal Pretraining}
Large-scale contrastive pretraining has enabled strong cross-modal alignment in a shared embedding space. Image-text models such as CLIP~\cite{radford2021learning} and SigLIP~\cite{Tschannen2025SigLIP2M,Zhai2023SigmoidLF} learn transferable visual representations, while audio-text models such as CLAP~\cite{laion-clap,m2d-clap,spatialclap} extend this paradigm to sound. More general encoders further bind multiple modalities: AudioCLIP~\cite{guzhov2021audioclip} aligns audio and vision through text, and ImageBind~\cite{girdhar2023imagebind} uses image-based alignment to unify image, text, audio, and other signals, supporting retrieval and zero-shot transfer. These encoders are effective semantic foundations for multimodal systems~\cite{Qwen-Audio,han2023onellm,panagopoulou2023xinstructblip}, and focus on emphasizing semantics (\emph{what} is present) and may omit location (\emph{where} entities are). Spatial cues are often weakly preserved: rotations and flips may be treated as semantic equivalents, and audio is frequently downmixed to monaural signals. SceneBind addresses this gap by preserving spatial factors while remaining compatible with alignment-based pretraining.

\subsection{Spatially Grounded Multimodal Representations}
Multimodal spatial reasoning requires estimating where entities are and how they relate in 3D, including direction, distance, and relative geometry~\cite{yang2024think,chen2025savvy}. Vision-centric approaches model 3D structure with point clouds~\cite{garg2024robohop}, graphs~\cite{conceptfusion,gu2024conceptgraphs,yang20243dmem3dscenememory}, voxel grids~\cite{wang2023gridmm}, maps~\cite{huang23vlmaps,yang2024think}, and neural fields~\cite{chen2024chatsplat,shi2024language3dgs}. Object-centric methods, including detection and grounding~\cite{ren2024grounded,Liu2023GroundingDM,Ren2024GroundingD1,detr,deformable-detr} and multi-modal scene graphs~\cite{Chatterjee2021VisualSGforaudiosep,avdynamicsSG}, further capture fine-grained scene structure, but are often local, foreground-focused, or not designed as unified cross-modal representations.

Audio is an important modality for identifying and spatially localizing objects and events that produce sound. However, many audiovisual spatial grounding use monaural audio and rely on visual cues to localize sound~\cite{mo2023audio,ye2024lavss,owens2018audio,hu2020discriminative,Tian2021CyclicCO}, making spatial inference challenging under occlusion, out-of-view sources, or dynamic scenes. Spatial audio provides direct physical cues, such as inter-channel phase and level differences, for estimating direction and distance~\cite{adavanne2018seldnet,diaz2020robust,zheng2024bat,spatialclap,elsa,dementyev2026phasecodermicrophonegeometryagnosticspatial,ryu2026hear}. When spatial audio was taken into account in prior works, it was often from synthetic data or constrained sensors, limiting generalization to noisy real-world environments. In contrast, SceneBind learns unified semantic-spatial representations from real-world audio, vision, and language, combining global semantics with semantic-spatial slots to capture holistic scene context and object-level alignment.

\section{Method}
\subsection{Problem Setup and System Overview}
\label{sec:problem_setup}
We define a scene as a 3D semantic–spatial entity that captures the context of \emph{what} is present and \emph{where} objects are located in the scene. A scene can be observed through multiple modalities, including image $I$, audio $A$, and text $T$. The text modality consists of a global scene description and a set of object-level clauses, each paired with spatial attributes $\mathbf{r}_k = (\theta_k, \phi_k, d_k)$ denoting camera-coordinate azimuth, elevation, and distance.
SceneBind maps each modality into a shared semantic–spatial representation $\mathcal{X}$ consisting of a global embedding and $K$ object slots
\begin{equation}
\mathcal{X} = \left( \mathbf{s}_{\text{global}}, \left\{ (\mathbf{s}_k, \mathbf{r}_k, c_k) \right\}_{k=1}^{K} \right),
\end{equation}
where $\mathbf{s}_{\text{global}} \in \mathbb{R}^d$ encodes scene-level semantics, and each object slot consists of a semantic embedding $\mathbf{s}_k \in \mathbb{R}^d$, spatial attributes $\mathbf{r}_k$, and a confidence score $c_k$.

\begin{figure}[t]
\centering
\includegraphics[width=1.0\linewidth]{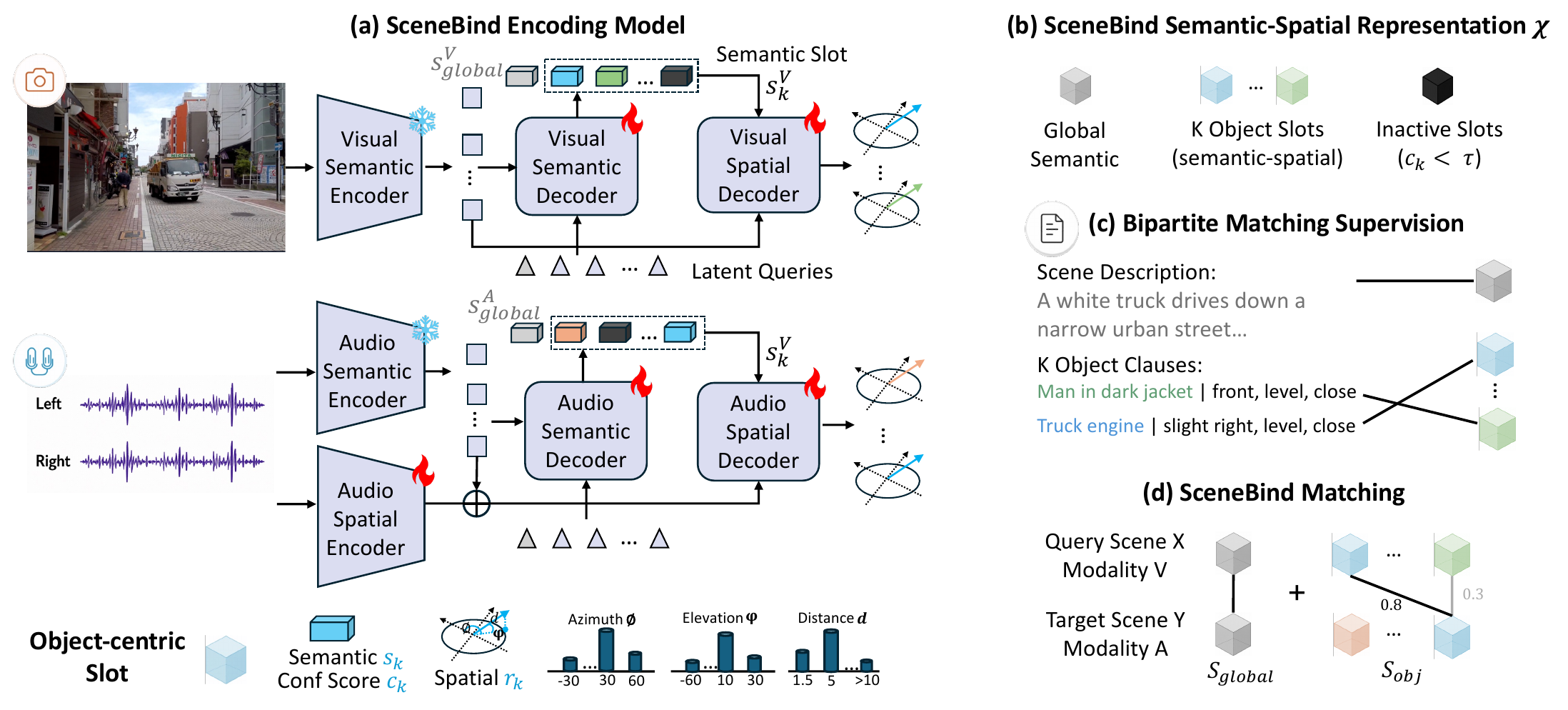}
\vspace{-3mm}
\caption{\textbf{SceneBind overview.}
(a) \textbf{SceneBind Encoding Model} maps multimodal inputs to a global embedding $\mathbf{s}^{\text{global}}$ and object-centric slots $(\mathbf{s}_k,\mathbf{r}_k,c_k)$.
(b) \textbf{SceneBind Representation} $\mathcal{X}$ represents each scene with a global embedding and $K$ object-centric semantic-spatial slots.
(c) Bipartite matching aligns slots with GT object clauses for supervision.
(d) \textbf{SceneBind Matching} combines global similarity $S_{global}$ with object slot alignment score $S_{obj}$ for cross-modal scene comparison.}
\label{fig:main_details}
\vspace{-4mm}
\end{figure}

As illustrated in Figure~\ref{fig:main_details}, SceneBind encodes each modality into a shared semantic-spatial representation $\mathcal{X}$ composed of a global embedding and object-centric semantic-spatial slots (Sec.~\ref{sec:encoding_model}). It is trained with global alignment and bipartite matching supervision (Sec.~\ref{sec:training}) to align global-level semantics across modalities and ground predicted slots to ground-truth (GT) text clauses. For inference, SceneBind Matching compares scene representations by combining global semantic similarity with object-centric slot alignment over $(\mathbf{s}_k, \mathbf{r}_k, c_k)$ (Sec.~\ref{sec:matching}).

\subsection{SceneBind Encoding Model}
\label{sec:encoding_model}

Given an input sample from modalities $m\in\{A{-Audio},V{-Visual}\}$, SceneBind uses modality-specific encoders and decoders to extract semantic--spatial context tokens, obtain global and object semantic states, and decode object spatial attributes. Together, these modules produce a scene representation $\mathcal{X}^m$.

\paragraph{Semantic-Spatial Context Tokens.}
For visual modality, we use the frozen visual tower of a vision-text pretrained encoder SigLIP2~\cite{Tschannen2025SigLIP2M} to produce patch tokens $\mathbf{F}^V\in\mathbb{R}^{L_V\times d}$ in the vision-text space, where the patch sequence preserves image layout and provides both semantic and spatial evidence.
For the audio modality, we use the frozen audio tower of M2D-CLAP~\cite{m2d-clap} to encode the monaural mixture into semantic patch tokens in the M2D-CLAP audio-text embedding space. Because this embedding space differs from the SigLIP2 vision-text space, we train a lightweight alignment module to map the M2D-CLAP audio tokens into the SigLIP2 embedding space and obtain $\mathbf{F}^{A}_{\mathrm{sem}}\in\mathbb{R}^{L_A\times d}$.

To capture spatial cues, we convert the binaural waveform into a four-channel time-frequency representation, consisting of left/right magnitude and phase, and feed it to a convolutional spatial encoder trained from scratch. The resulting spatial features are projected to dimension $d$, interpolated to length $L_A$, and added to $\mathbf{F}^{A}_{\mathrm{sem}}$ to form the audio context $\mathbf{F}^{A}\in\mathbb{R}^{L_A\times d}$.

\paragraph{Semantic Decoding.}
SceneBind uses a cross-attention semantic decoder $D^m_{\mathrm{sem}}$ to extract global and object semantics from visual tokens $\mathbf{F}^V$ for vision and audio semantic tokens $\mathbf{F}^{A}_{\mathrm{sem}}$ for audio. 
A learned global query aggregates holistic scene content, analogous to pooling in pretrained encoders~\cite{Tschannen2025SigLIP2M,m2d-clap}, and outputs a global embedding $\mathbf{s}^{m}_{\mathrm{global}}$. Distinctively, SceneBind introduces $K$ learned latent object queries that act as semantic anchors, decoding object semantic states $\{\mathbf{h}^{m}_{k}\}_{k=1}^{K}$ from context tokens. Each object state is decoded to a semantic slot $\mathbf{s}^{m}_{k}$  and activity confidence $c^{m}_{k}$
\begin{equation}
\small
\mathbf{s}^{m}_{k}=g^m_{\mathrm{sem}}(\mathbf{h}^{m}_{k}),
\qquad
c^{m}_{k}=\sigma(g^m_{\mathrm{conf}}(\mathbf{h}^{m}_{k})).
\end{equation}

\paragraph{Spatial Decoding.}
The spatial decoder $D^m_{\mathrm{spa}}$ uses each object semantic slot $\mathbf{s}^{m}_{k}$ from $D^m_{\mathrm{sem}}$ as a query to decode spatial attributes $\hat{\mathbf{r}}^{m}_{k}=(\hat{\theta}^{m}_{k},\hat{\phi}^{m}_{k},\hat{d}^{m}_{k})$ from semantic-spatial context tokens. It cross-attends to visual patch tokens $\mathbf{F}^V$ for visual and for fused audio context $\mathbf{F}^{A}$, where $\mathbf{F}^{A}$ combines semantic audio tokens with binaural spatial cues. The decoder outputs $\mathbf{z}^{m}_{k}$ which is mapped by modality specific projection heads into spatial embeddings $\mathbf{u}^{m,\theta}_{k}$, $\mathbf{u}^{m,\phi}_{k}$ and $\mathbf{u}^{m,d}_{k}$. These are passed to shared azimuth, elevation, and distance classifiers. The final prediction $\hat{\mathbf{r}}^{m}_{k}$ is obtained by taking the argmax over logits of each classifier.

For text, the frozen text encoder~\cite{Tschannen2025SigLIP2M} produces a global scene embedding from the scene description and object-centric semantic embeddings from individual clauses. The spatial labels attached to object clauses are used directly as azimuth, elevation, and distance targets.

\subsection{Bipartite Matching Supervision}
\label{sec:training}
This section describes how we supervise the SceneBind representation, which consists of a global semantic embedding and object semantic-spatial slots. For global semantic supervision, we apply a symmetric InfoNCE~\cite{infonce} loss $\mathcal{L}_{\mathrm{global}}$ to paired samples of audio-visual, audio-text and visual-text, using scene descriptions as text targets. For each modality pair, samples originating from the same scene form positive pairs, while samples from different scenes within the minibatch form negative pairs. This follows standard contrastive alignment, pulling matched samples together and pushing unmatched samples apart in the shared space.

Object-centric supervision is challenging because the predicted slots are unordered and the number of annotated objects varies across scenes. 
To accommodate this variability, SceneBind predicts a fixed set of $K$ slots and assigns a subset of them to the available object annotations. Specifically, as shown in Figure~\ref{fig:main_details} (c), we use bipartite matching~\cite{hungarian} to assign slots to ground-truth objects before applying object-level losses. Each ground-truth object $j$ includes a semantic embedding $\mathbf{s}^{*}_j$ and spatial attribute $\mathbf{r}^{*}_j=(\theta^{*}_j,\phi^{*}_j,d^{*}_j)$. Thus for predicted slot $i$, the matching score is
\begin{equation}
\small
a_{ij}=\big[(1-\lambda)\cos(\mathbf{s}_i,\mathbf{s}^{*}_j)+\lambda\,\rho_{ij}\big](\alpha+\beta c_i).
\label{eq:matching_score}
\end{equation}
Here $\rho_{ij}$ denotes the spatial matching score, computed as the average softmax probability assigned by slot $i$ to the ground-truth azimuth, elevation, and distance bins of object $j$. The one-to-one assignment that maximizes the total matching score provides the slot supervision.

\paragraph{Object Grounding Supervision.}
The object grounding loss $\mathcal{L}_{\mathrm{obj}}$ grounds each matched slot to its paired object clause in both semantics and space.
We align the slot embedding to the clause embedding with cosine distance, and supervise azimuth, elevation, and distance using Gaussian-smoothed cross-entropy over discretized bins, so nearby spatial bins incur smaller penalties than distant ones, preserving local spatial continuity. Slot confidence is trained with binary cross-entropy, treating matched slots as positives and unmatched slots as negatives.

\paragraph{Object-Centric Contrastive Learning.}
Beyond grounding, the matched assignments define cross-modal object correspondences. This contrastive supervision encourages slots that refer to the same object to align across modalities, while keeping objects from different instances separable. We use an intra-scene loss $\mathcal{L}_{\mathrm{inst}}$ and a cross-scene loss $\mathcal{L}_{\mathrm{batch}}$.
The intra-scene loss $\mathcal{L}_{\mathrm{inst}}$ contrasts slots matched to the same ground-truth object across modalities as positives, and other slots within the same scene as negatives, encouraging separation of co-occurring objects. The cross-scene loss $\mathcal{L}_{\mathrm{batch}}$ contrasts matched objects across the minibatch, further improving object discriminability. For audio-visual pairs, we apply cross-scene contrastive learning to both semantic and spatial embeddings
\begin{equation}
\small
\mathcal{L}_{\mathrm{ctr}}(U,V)
=
-\log
\frac{\exp(\mathrm{sim}(u_i,v_i)/\tau)}
{\sum_j \exp(\mathrm{sim}(u_i,v_j)/\tau)} ,
\end{equation}
\begin{equation}
\small
\mathcal{L}_{\mathrm{batch}}^{A,V}
=
\mathcal{L}_{\mathrm{ctr}}(\mathbf{s}^{A}, \mathbf{s}^{V})
+
\sum_{q\in\{\theta,\phi,d\}}
\gamma_q\,
\mathcal{L}_{\mathrm{ctr}}(\mathbf{u}^{A,q}, \mathbf{u}^{V,q}),
\end{equation}
where $\gamma_q$ scales the contribution of each spatial axis, and matched audio-visual slots assigned to the same ground-truth clause are positives. For audio-text and visual-text pairs, $\mathcal{L}_{\mathrm{batch}}$ uses only semantic embeddings because text has no learned spatial-head features.

\paragraph{Overall Training Objective.}
SceneBind is trained with a combination of global-level alignment ($\mathcal{L}_{\mathrm{global}}$), object grounding ($\mathcal{L}_{\mathrm{obj}}$) and object-centric contrastive objectives ($\mathcal{L}_{\mathrm{inst}}, \mathcal{L}_{\mathrm{batch}}$).

We train in two stages. The first stage uses mixed supervision from AudioCaps~\cite{audiocaps}, MS-COCO~\cite{lin2014microsoft}, and our curated Binaural dataset with aligned spatial labels, combining broad semantic alignment with spatially grounded object-centric learning. The second stage further fine-tunes on the Binaural dataset only, specializing SceneBind for cross-modal semantic-spatial alignment at both global and object levels. Full training details are provided in the appendix.

\subsection{SceneBind Matching}
\label{sec:matching}
At inference time, we propose a unified matching scheme (Figure~\ref{fig:main_details} (d)) that compares scenes using both global semantic alignment and object-centric semantic-spatial correspondence. Given a query scene $\mathcal{X}^q$ and a candidate scene $\mathcal{X}^c$, the global score is the cosine similarity between their scene-level embeddings, denoted as $S_{\mathrm{global}}$.
For object-centric matching, we first keep active slots according to their confidence scores. Each query slot retrieves its best-matching candidate slot under a semantic-spatial score weighted by both slot confidences
\begin{equation}
\small
S_{\mathrm{obj}}
=
\frac{
\sum_i w_{i j_i^*}\,
\cos(\mathbf{s}^{q}_{i}, \mathbf{s}^{c}_{j_i^*})\,\rho_{i j_i^*}
}{
\sum_i w_{i j_i^*}+\epsilon
},
\quad
j_i^*=\arg\max_j
w_{ij}\cos(\mathbf{s}^{q}_{i}, \mathbf{s}^{c}_{j})\rho_{ij},
\quad
w_{ij}=\sqrt{c_i^q c_j^c}.
\end{equation}
Here $\rho_{ij}$ measures spatial similarity between predicted azimuth, elevation, and distance distributions, and $w_{ij}$ is the geometric-mean confidence, which gives higher weight to reliable slot pairs. Each query slot selects its best candidate independently.
Finally, SceneBind combines global and object scores after query wise normalization:
$
\small
S(\mathcal{X}^q,\mathcal{X}^c)
=
\mathrm{norm}(S_{\mathrm{global}})
+
\lambda_{\mathrm{match}}\,
\mathrm{norm}(S_{\mathrm{obj}}),
$
where $\lambda_{\mathrm{match}}$ balances scene-level semantic alignment and object-level semantic-spatial matching.

\section{Experiments}
\subsection{Spatially Grounded In-the-Wild Data Curation}
\label{sec:data-curation}
We curate spatially grounded omni-modal data from two sources. The primary \emph{Binaural} corpus is collected from 21{,}927 in-the-wild videos with binaural audio, providing real spatial audio cues. Our multi-stage pipeline filters source videos, annotates candidate events, verifies cross modal consistency, balances spatial and semantic coverage, and applies human review for evaluation. We use Gemini~\cite{comanici2025gemini} to propose short events from video clips, linking audible events to visible entities and estimating camera-coordinate spatial labels from visual evidence. The proposals are then verified with pretrained audio and visual encoders~\cite{girdhar2023imagebind,laion-clap,m2d-clap} and balanced to reduce spatial and semantic redundancy.
Each retained clip contains a two-second audio-image pair, a scene description, and object clauses with instance semantic description and spatial labels (azimuth, elevation, and distance). The final training split contains 38{,}430 clips with 207{,}836 objects, and the human-verified Binaural benchmark contains 1{,}066 clips. We also build a \emph{Sphere360} benchmark from $360^\circ$ ambisonic videos~\cite{Liu2025OmniAudioGS} for zero-shot evaluation under full-sphere viewpoint variation. More details are provided in the appendix.

\subsection{Tasks and Evaluation}
\label{sec:tasks_eval}

We evaluate SceneBind to answer: (1) How does SceneBind improve over global semantics-only representations across retrieval and object grounding tasks? (2) How do training and inference designs affect semantic-spatial alignment? (3) How does SceneBind enable zero-shot generalization?

\textbf{Cross-modal Scene Retrieval.}
We evaluate retrieval across Audio (A), Visual (V) and Text (T) modalities (Table~\ref{tab:retrieval_scene_spatial}), where each scene includes binaural audio, an image, and a text description (global caption + object clauses). Given a query in one modality (or a fused pair), the task is to retrieve the corresponding scene in another modality. We report bidirectional Recall@1 (R@1), i.e., the fraction of queries whose correct match is ranked first, for $A\leftrightarrow V$, $A\leftrightarrow T$ and $V\leftrightarrow T$, test sample pool size 1046.
We further evaluate zero-shot generalization on Sphere360 (Table~\ref{tab:sphere_hard_scene}), where each text query uses a single dominant object clause with the global caption and candidates are spatially augmented views of the same scene, requiring fine-grained spatial discrimination.

\textbf{Spatial Retrieval.}
We evaluate spatial retrieval using object text clauses with spatial attributes (Table~\ref{tab:retrieval_scene_spatial}). Candidates are ranked by the maximum spatial similarity over object-centric predictions within each sample. We report spatial mAP, which measures how highly candidates with the correct spatial label are ranked, and spatial R@1, computed within the same semantic cluster but different spatial configurations to test spatial discrimination.

\textbf{Object Grounding.}
Object grounding evaluates whether the model can correctly localize a queried object in space. Given an object query, we encode its semantic description (without spatial attributes) and match it to predicted object slots based on semantic similarity and confidence (Table~\ref{tab:grounding}). The best-matching slot is selected for evaluation. We report recall as the fraction of queries with semantic similarity $>0.5$, and accuracy as those additionally with all spatial labels (azimuth, elevation, distance) correct. We also report per-attribute spatial accuracy.

\textbf{Zero-shot Downstream Tasks.}
We evaluate zero-shot transfer to multimodal tasks requiring joint spatial and semantic reasoning via egocentric audio-visual localization~\cite{huang2023egocentric} (Table~\ref{tab:avloc}). Given stereo audio and an image, the goal is to predict the sounding region in the image. We directly apply SceneBind without task-specific training. We report \textit{cIoU} at multiple thresholds (e.g., 0.2, 0.3, 0.4), measuring overlap between predicted and GT, and \textit{AUC} summarizing performance across thresholds.

\subsection{Baselines}
\label{sec:baselines}
We compare with representative multimodal encoders, including AudioCLIP~\cite{guzhov2021audioclip}, CLAP variants (LAION-CLAP~\cite{laion-clap}, M2D-CLAP~\cite{m2d-clap}), SpatialCLAP~\cite{spatialclap}, and ImageBind~\cite{girdhar2023imagebind}. We pair CLAP-based audio encoders with SigLIP2~\cite{Tschannen2025SigLIP2M}, while AudioCLIP and ImageBind use native encoders. For zero-shot evaluation, we use native text encoders and chained scoring for AV retrieval. As these models produce a single embedding per sample, we treat them as a single object-centric prediction (confidence 1.0, no spatial attributes), match each clause independently, and aggregate similarities for retrieval.
For further fair comparison, we fine-tune a selected set of strong and representative baselines, including ImageBind*, M2D-CLAP* paired with SigLIP2*, and SpatialCLAP* paired with SigLIP2* (which supports spatial audio input), under the same data and training setup as SceneBind by learning projection heads into a shared space aligned with frozen SigLIP2 text features, using global and multi-positive (object clause) contrastive losses. Additional details are provided in the appendix.

\begin{table*}[t]
\centering
\setlength{\tabcolsep}{4.5pt}
\caption{\textbf{Cross-modal scene retrieval (R@1) and object-level spatial retrieval} on unseen Binaural test set. SceneBind outperforms all baselines, with largest gains on text-conditioned retrieval (especially V$\leftrightarrow$T) and spatial metrics, demonstrating effective semantic–spatial representation for scenes.}

\label{tab:retrieval_scene_spatial}
\begin{tabular}{ll|cccc|ccc}
\toprule
\multicolumn{2}{c|}{Encoder} & \multicolumn{4}{c|}{Scene Retrieval (R@1 $\uparrow$)}  & \multicolumn{3}{c}{Spatial Retrieval ($\uparrow$)} \\
Audio--Text & Visual--Text & A$\leftrightarrow$V  & A$\leftrightarrow$T & V$\leftrightarrow$T & Avg  & R@1 & mAP & Avg \\
\midrule
AudioCLIP~\cite{guzhov2021audioclip} & AudioCLIP~\cite{guzhov2021audioclip} & 0.3 & 0.3 & 20.6 & 7.1 & 9.3 & 28.1 & 18.7 \\
SpatialCLAP~\cite{spatialclap} & SigLIP2~\cite{Tschannen2025SigLIP2M} & 0.4 & 1.9 & \underline{51.2} & 17.8 & 10.1 & 28.0 & 19.0 \\
LAION-CLAP~\cite{laion-clap} & SigLIP2~\cite{Tschannen2025SigLIP2M} & 0.5 & 5.9 & \underline{51.2} & 19.2 & 10.1 & 28.7 & 19.4 \\
M2D-CLAP~\cite{m2d-clap} & SigLIP2~\cite{Tschannen2025SigLIP2M} & 0.3 & 9.0 & \underline{51.2} & 20.2 & \underline{11.4} & \underline{29.6} & \underline{20.5} \\
ImageBind~\cite{girdhar2023imagebind} & ImageBind~\cite{girdhar2023imagebind} & 7.8 & 7.6 & 48.0 & 21.2 & 10.6 & 28.7 & 19.6 \\
\midrule
M2D-CLAP$^{*}$~\cite{m2d-clap} & SigLIP2$^{*}$~\cite{Tschannen2025SigLIP2M} & \underline{19.3} & \underline{11.2} & 42.8 & \underline{24.4} & 10.1 & 28.9 & 19.5 \\
ImageBind$^{*}$~\cite{girdhar2023imagebind} & ImageBind$^{*}$~\cite{girdhar2023imagebind} & \underline{19.3} & 10.9 & 35.4 & 21.9 & 10.4 & 29.4 & 19.9 \\
SpatialCLAP$^{*}$~\cite{spatialclap} & SigLIP2$^{*}$~\cite{Tschannen2025SigLIP2M} & 16.1 & 9.3 & 44.2 & 23.2 & 10.6 & 29.2 & 19.9 \\
\bf SceneBind (Ours) & \bf SceneBind (Ours) & \textbf{21.8} & \textbf{17.0} & \textbf{65.3} & \textbf{34.7} & \textbf{28.9} & \textbf{48.0} & \textbf{38.4} \\
\bottomrule
\end{tabular}
\vspace{-5mm}
\end{table*}

\subsection{\textbf{Main Results}}

\textbf{Cross-modal Scene Retrieval.}
SceneBind achieves the best overall performance across A$\leftrightarrow$V, A$\leftrightarrow$T, and V$\leftrightarrow$T retrieval tasks (Table~\ref{tab:retrieval_scene_spatial}). Compared to pretrained encoders, fine-tuning (*) improves A$\leftrightarrow$V retrieval and A$\leftrightarrow$T moderately, but yields limited or negative gains on V$\leftrightarrow$T. In contrast, SceneBind consistently outperforms all baselines, achieving 65.3 on V$\leftrightarrow$T (+28\% over pretrained, +48\% over finetuned) and 17.0 on A$\leftrightarrow$T (+52\%), with the largest gains on text-relevant retrieval—especially V$\leftrightarrow$T. This highlights that joint semantic-spatial modeling with object-centric matching of SceneBind improves scene retrieval beyond global semantic encoding.

On the Sphere360 hard pool (Table~\ref{tab:sphere_hard_scene}), where candidates share the same semantics but differ spatially, fine-tuning provides little improvement, while SceneBind achieves 29.3 (+56\% over best finetuned), demonstrating strong zero-shot spatial discrimination under semantic ambiguity for scene retrieval.

\textbf{Spatial Retrieval.}
SceneBind significantly outperforms all baselines on spatial retrieval (Table~\ref{tab:retrieval_scene_spatial}), achieving 28.9 R@1 and 48.0 mAP, compared to the best baseline, corresponding to +153\% and +62\% improvements, respectively. Notably, fine-tuning existing encoders yields minimal gains, since they mostly rely on global semantics and lack explicit spatial grounding. In contrast, SceneBind’s object-centric spatial representation and semantic–spatial matching enable accurate retrieval within semantically identical but spatially varying candidates.

\textbf{Object Grounding.}
SceneBind enables text-guided object grounding via its object-centric semantic-spatial representation. As shown in Table~\ref{tab:grounding}, SceneBind achieves overall grounding accuracies of 20.1\% using Audio (A) and 19.1\% using Visual (V). More importantly, its spatial attribute accuracies are substantially above random chance, including 83.2\% for audio elevation recovery and 67.1\% and 66.1\% for visual left/right and distance recovery, respectively. Audio provides stronger elevation recovery, whereas vision performs better on left/right azimuth and distance estimation. Together, these results demonstrate that SceneBind learns meaningful associations between object text semantics and spatial attributes across both modalities.

\subsection{Ablations}

\textbf{Training Objectives Ablations.}
Removing object-level semantic supervision (no $\mathcal{L}_{\text{sem}}$) degrades grounding (19.6$\rightarrow$8.8) and spatial retrieval (38.4$\rightarrow$35.2), as object representations are no longer aligned with text queries, while scene retrieval slightly increases (34.7$\rightarrow$36.0) due to global bias. Removing intra scene contrastive loss ($\mathcal{L}_{\text{inst}}$) leads to overall degradation (Avg 30.9$\rightarrow$27.2), with notable drops in spatial (38.4$\rightarrow$31.1) and grounding (19.6$\rightarrow$16.1), indicating weaker object discrimination. Removing batch level contrastive loss ($\mathcal{L}_{\text{batch}}$) primarily affects spatial retrieval (38.4$\rightarrow$36.5), as it encodes spatial discrimination across scenes, while intra scene alignment still preserves semantic matching. Combining all losses yields the best overall performance.

\textbf{Data Recipe with Two-Stage Training.}
We train SceneBind with multiple source of data, including our novel Binaural dataset (spatial AVT), AudioCaps~\cite{audiocaps} (AT), and MS-COCO~\cite{lin2014microsoft} (VT). Using Binaural only yields lower performance due to weaker semantic alignment. Training all data in a single stage improves scene (34.7$\rightarrow$35.2) and grounding (19.6$\rightarrow$19.9) but degrades spatial retrieval (38.4$\rightarrow$36.5), as non spatial data dilutes spatial supervision. A two-stage schedule in which we train on all data for global alignment and then fine tune on Binaural data for better spatial grounding achieves the best balance, preserving spatial performance while improving overall alignment.

\textbf{Matching Policy Justification.}
Global matching captures holistic scene semantics and is strong for semantic retrieval, but lacks spatial precision. Adding slots improves all metrics as shown in Table~\ref{tab:match_policy}, especially Spatial retrieval (25.4 to 38.4), showing the need for object-centric slots which provide the missing spatial location signal. Slot-only matching preserves Spatial performance but loses scene context, causing large scene retrieval accuracy drops. The slot-only ablations further justify our \textit{Best match} design. In particular, Hungarian one-to-one assignment improves T$\rightarrow$A/V when clean GT clauses query predicts slots, but degrades A/V$\rightarrow$T by forcing noisy predicted slots to match GT clauses. Removing confidence mainly degrades prediction-to-prediction (A$\leftrightarrow$V) and prediction-to-GT (A/V$\rightarrow$T) matching since noisy low-quality slot pairs are no longer down-weighted.

\textbf{Object-Centric Slot Scaling.}
We analyze how the number of object-centric slots affects semantic-spatial performance (Fig.~\ref{fig:slot_grounding}). Beyond a moderate number of slots (i.e., $>10$), performance saturates, suggesting that a small set of $\sim$10 is sufficient to capture scene-level structure. Visual tasks benefit from more slots (peaking around 25) due to higher object density, while audio which typically has fewer concurrent events peaks with fewer slots (around 5). For zero-shot audio-visual localization (Sec.~\ref{sec:avloc}), more slots provide better candidate regions and improve coverage.

\begin{figure*}[t]
\centering
\small
\captionsetup{
  font=small,
  labelfont=bf,
  justification=raggedright,
  singlelinecheck=false,
  skip=3pt
}

\begin{minipage}[t]{0.45\textwidth}
\centering
\vspace{0pt}

\captionof{table}{\textbf{Object grounding with per-attribute spatial accuracy.}
SceneBind enables consistent semantic--spatial grounding for A/V objects.}
\label{tab:grounding}
\setlength{\tabcolsep}{2.5pt}
\begin{tabular*}{\linewidth}{@{\extracolsep{\fill}}lccccc@{}}
\toprule
Modality & Acc $\uparrow$  & Azi $\uparrow$  & L/R $\uparrow$  & Elev $\uparrow$  & Dist $\uparrow$  \\
\midrule
A & \bf 20.1 & \bf 39.4 & 62.7 & \bf 83.2 & 58.4 \\
V & 19.1 & 35.5 & \bf 67.1 & 76.1 & \bf 66.1 \\
\bottomrule
\end{tabular*}

\vspace{7pt}

\captionof{table}{\textbf{SceneBind Matching ablations.}
Global and slot matching are complementary; slot-only ablations support our design (\textit{Best match}). For all metrics, the higher, the better.}
\label{tab:match_policy}

\setlength{\tabcolsep}{2.3pt}
\begin{tabular*}{\linewidth}{@{\extracolsep{\fill}}lcccc@{}}
\toprule
Strategy & AV & T$\rightarrow$A/V & A/V$\rightarrow$T & Spatial \\
\midrule
Global & 21.3 & 36.1 & 34.2 & 25.4 \\
\bf Global+Slot & \bf 21.8 & \bf 42.4 & \bf 39.9 & \bf 38.4 \\
\midrule
\multicolumn{5}{@{}l}{Slot-only:} \\
Best match & 11.5 & 18.6 & 17.3 & 38.4 \\
Hungarian & 10.9 & 21.1 & 6.9 & 38.4 \\
w/o conf & 9.5 & 22.8 & 15.7 & 38.4 \\
\bottomrule
\end{tabular*}

\end{minipage}
\hfill
\begin{minipage}[t]{0.51\textwidth}
\centering
\vspace{0pt}

\captionof{table}{\textbf{Training objectives ablations.}
Object semantic loss is key for grounding, intra scene contrastive loss
for discrimination, and batch level semantic-spatial contrastive loss for spatial retrieval.}
\setlength{\tabcolsep}{3.0pt}
\begin{tabular*}{0.9\linewidth}{@{\extracolsep{\fill}}ccc|ccc@{}}
\toprule
$\mathcal{L}_{\text{sem}}$ &
$\mathcal{L}_{\text{inst}}$ &
$\mathcal{L}_{\text{batch}}$ &
Scene $\uparrow$ & Spatial $\uparrow$  & Ground $\uparrow$  \\
\midrule
\checkmark & \checkmark &  & {35.5} & {36.5} & \textbf{19.6} \\
\checkmark &  & \checkmark & 34.3 & 31.1 & 16.1 \\
 & \checkmark & \checkmark & \textbf{36.0} & 35.2 & 8.8 \\
\checkmark & \checkmark & \checkmark & 34.7 & \textbf{38.4} & \textbf{19.6} \\
\bottomrule
\end{tabular*}

\vspace{8pt}

\captionof{table}{\textbf{Training schedule and data.}
Two-stage training (All $\rightarrow$ Binaural) preserves spatial performance
while improving overall alignment.}
\setlength{\tabcolsep}{3.0pt}
\begin{tabular*}{0.93\linewidth}{@{\extracolsep{\fill}}ll|ccc@{}}
\toprule
Stage 1 & Stage 2 & Scene $\uparrow$  & Spatial $\uparrow$  & Ground $\uparrow$  \\
\midrule
Binaural & -- & {33.8} & 35.2 & {19.5} \\
All & Binaural & 34.7 & \textbf{38.4} & 19.6 \\
All & All & \textbf{35.2} & {36.5} & \textbf{19.9} \\
\bottomrule
\end{tabular*}

\end{minipage}

\end{figure*}

\begin{table*}[t]
\centering
\small

\begin{minipage}[t]{0.38\linewidth}
\centering
\vspace{0pt}

\includegraphics[width=\linewidth]{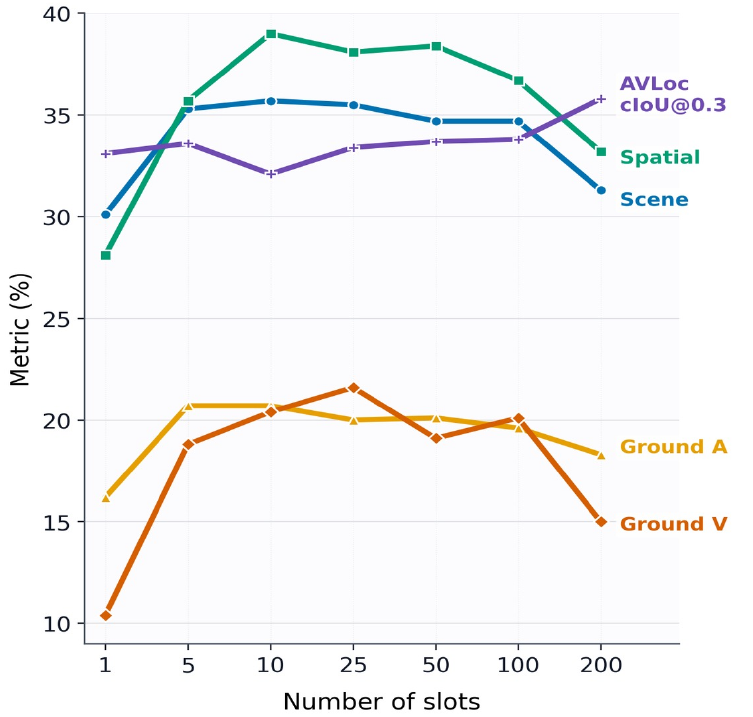}

\vspace{-1mm}
\captionof{figure}{\textbf{Object slot scaling.} Performance saturates or drops beyond a moderate number; $\sim$10 slots suffice.}
\label{fig:slot_grounding}

\end{minipage}
\hfill
\begin{minipage}[t]{0.60\linewidth}
\centering
\vspace{0pt}

\caption{\textbf{Zero-shot Sphere360 hard-pool retrieval.} 
Scene-level retrieval under identical semantics but varying spatial configurations. SceneBind significantly improves performance, demonstrating strong spatial discrimination under semantic ambiguity. For all metrics, the higher, the better.}
\label{tab:sphere_hard_scene}

\setlength{\tabcolsep}{4pt}
\renewcommand{\arraystretch}{0.95}

\begin{tabular}{ll|cccc}
\toprule
Audio--Text & Visual--Text & AV  & VT & AT & Avg \\
\midrule
AudioCLIP~\cite{guzhov2021audioclip} & AudioCLIP~\cite{guzhov2021audioclip} & 11.2 & 16.0 & 4.9 & 10.7 \\
SpatialCLAP~\cite{spatialclap} & SigLIP2~\cite{Tschannen2025SigLIP2M} & 13.6 & 19.9 & 1.9 & 11.8 \\
LAION-CLAP~\cite{laion-clap} & SigLIP2~\cite{Tschannen2025SigLIP2M} & 17.5 & 19.9 & 8.7 & 15.4 \\
M2D-CLAP~\cite{m2d-clap} & SigLIP2~\cite{Tschannen2025SigLIP2M} & 18.4 & 19.9 & \underline{10.2} & 16.2 \\
ImageBind~\cite{girdhar2023imagebind} & ImageBind~\cite{girdhar2023imagebind} & 15.0 & 18.0 & 5.3 & 12.8 \\
\midrule
M2D-CLAP$^{*}$~\cite{m2d-clap} & SigLIP2$^{*}$~\cite{Tschannen2025SigLIP2M} & 19.9 & 20.4 & 8.3 & 16.2 \\
ImageBind$^{*}$~\cite{girdhar2023imagebind} & ImageBind$^{*}$~\cite{girdhar2023imagebind} & \underline{23.3} & \underline{22.8} & \underline{10.2} & \underline{18.8} \\
SpatialCLAP$^{*}$~\cite{spatialclap} & SigLIP2$^{*}$~\cite{Tschannen2025SigLIP2M} & 19.4 & 19.9 & 9.7 & 16.3 \\
\bf SceneBind & \bf SceneBind & \textbf{25.7} & \textbf{32.0} & \textbf{30.1} & \textbf{29.3} \\
\bottomrule
\end{tabular}

\end{minipage}
\vspace{-5mm}
\end{table*}

\begin{figure}[htb]
    \centering
    \includegraphics[width=1.0\linewidth]{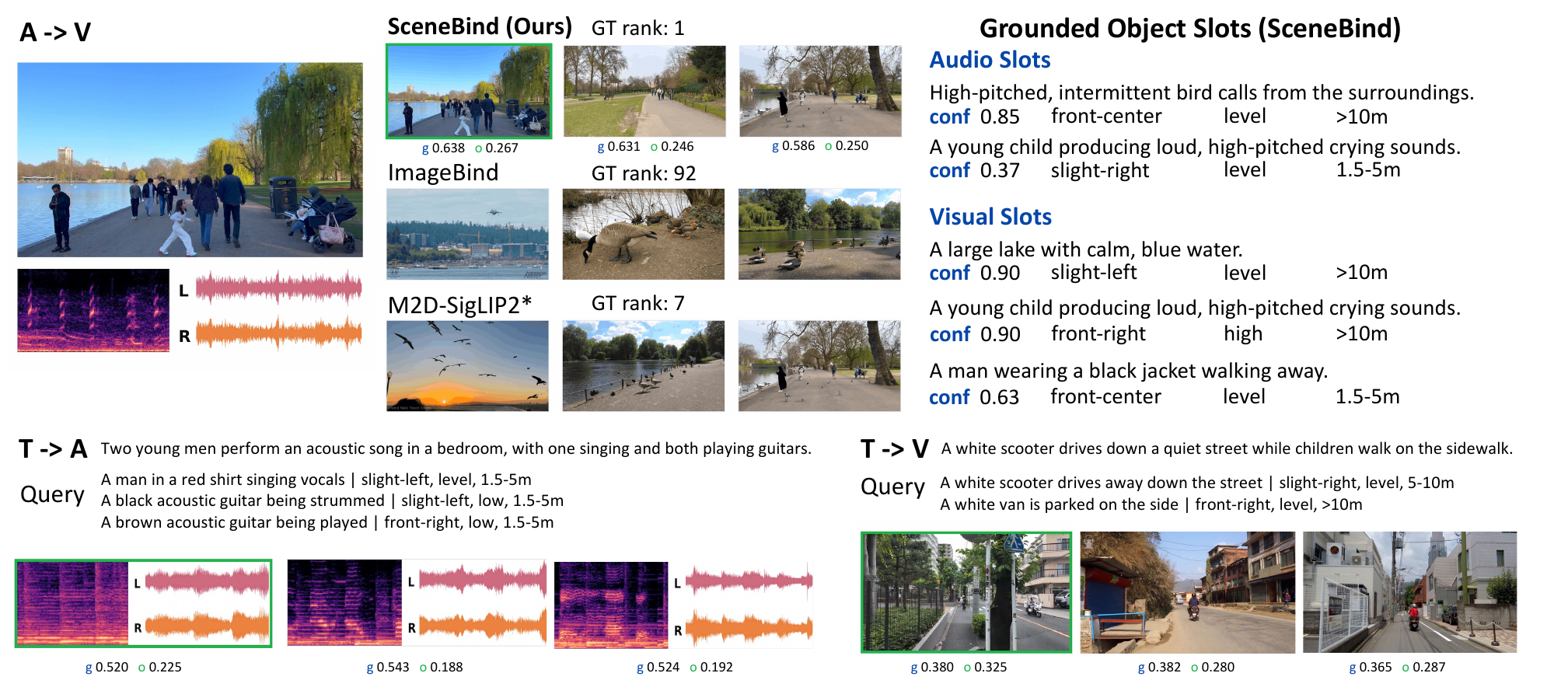}
    \vspace{-3mm}
\caption{\textbf{Qualitative results.}
SceneBind retrieves semantically and spatially aligned scenes across audio, vision, and text. We show top-3 candidates with global similarity ($g$), object-centric matching score ($o$), and grounded slots with confidence and spatial labels.}

    \label{fig:qualitative}
\end{figure}

\begin{table*}[t]
\centering
\small

\begin{minipage}[t]{0.48\textwidth}
\centering
\setlength{\tabcolsep}{4pt}
\renewcommand{\arraystretch}{0.95}
\captionof{table}{\textbf{Ego AV localization.} SceneBind (AV) uses binaural-only training, while SceneBind uses full multi-source training; both are zero-shot and outperform prior finetuned methods, with full training further improving performance.}

\label{tab:avloc}
\begin{tabular}{l|ccc|c}
\toprule
 & \multicolumn{3}{c|}{cIoU $\uparrow$ } & AUC $\uparrow$  \\
Method & @0.2 & @0.3 & @0.4 &  \\
\midrule
Attention~\cite{senocak2018learning} & 7.12 & -- & -- & 6.42 \\
STM~\cite{li2021space}        & 12.10 & 7.64 & 4.01 & 8.87 \\
Hardway~\cite{chen2021localizing}    & 24.51 & 13.55 & 6.10 & 13.38 \\
SSPL~\cite{song2022self}           & 13.62 & 8.10 & 4.45 & 9.56 \\
Mix~\cite{hu2022mix}    & 26.01 & 15.25 & 9.90 & 15.39 \\
SSS~\cite{ryu2025seeing}    & 9.76 & 9.89 & 10.10 & 7.50 \\
AVLoc~\cite{huang2023egocentric}           & 38.71 & 19.42 & 10.51 & 18.38 \\
\midrule
\bf SceneBind (AV) & \underline{43.94} & \underline{26.34} & \underline{13.91} & \underline{20.80} \\
\bf SceneBind        & \textbf{52.65} & \textbf{33.73} & \textbf{19.47} & \textbf{24.39} \\
\bottomrule
\end{tabular}

\end{minipage}
\hfill
\begin{minipage}[t]{0.48\textwidth}
\centering
\vspace{0pt}
\includegraphics[width=\linewidth]{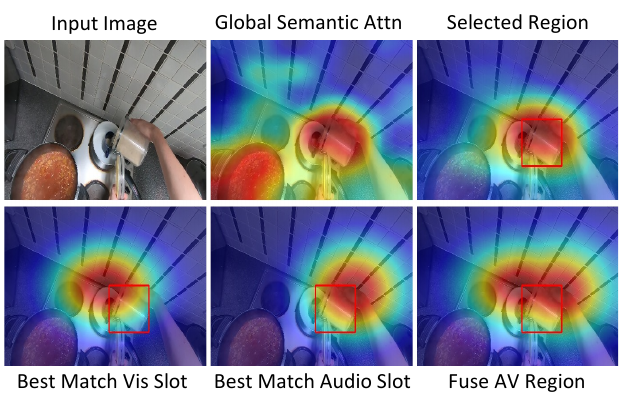}

\captionof{figure}{\textbf{AV localization visualization.} Global semantic attention proposes sounding regions, refined by spatial consistency of SceneBind slots.}
\label{fig:av_loc_attn}
\end{minipage}
\vspace{-5mm}
\end{table*}

\subsection{Qualitative Results}
\label{sec:qual}
Figure~\ref{fig:qualitative} compares SceneBind with the strongest zero-shot baseline, ImageBind, and the strongest fine-tuned baseline, M2D-SigLIP2*. In A$\rightarrow$V retrieval, SceneBind ranks the GT park scene first and retrieves candidates with similar layouts, while the baselines capture coarse semantics but miss key spatial cues such as ``a child crying on the right''. SceneBind Matching identifies cues such as ``bird sounds'' and ``a child crying on the right'' from object-slot predictions, while the object score ($o$) complements the global score ($g$). These balance matches of spatial objects with overall scene context.  In T$\rightarrow$A and T$\rightarrow$V retrieval, SceneBind also matches queried objects with their spatial attributes; in the scooter example, the correct image is selected by a higher slot score despite a lower global score. These examples show SceneBind's ability to align both semantics and locations.

\subsection{Zero-shot Downstream Application: Egocentric Spatial Audio-Visual Localization}
\label{sec:avloc}
SceneBind generalizes to egocentric AV localization without task-specific training (Table~\ref{tab:avloc}), achieving strong cIoU and AUC, demonstrating effective transfer of semantic-spatial representations.

As shown in Fig.~\ref{fig:av_loc_attn}, the global semantic query induces an attention map over visual patches, serving as a region-of-interest prior.
We perform cross-modal matching between audio and visual object-centric slots using semantic and spatial consistency, and select the best pair. The predicted spatial attributes (azimuth, elevation) define a Gaussian spatial prior, which is fused with the attention map to refine localization, yielding the final sounding region. More details are described in the appendix.

\section{Conclusion}

We presented \textbf{SceneBind}, a unified omni-modal representation for joint semantic and spatial representation across vision, audio and language. By combining global semantics with object semantic-spatial slots, SceneBind models \emph{what} and \emph{where}, enabling spatial-semantic alignment for cross modal retrieval and grounding. We introduce a curated spatially aligned binaural dataset, a joint semantic-spatial training protocol, and a unified matching policy integrating global and object representations. SceneBind achieves strong performance along with effective zero-shot generalization.

\textbf{Limitations and Future Works.}
SceneBind may benefit from scaling real world spatially aligned multimodal data with richer in-the-wild annotations, and from extending to longer temporal windows to model object motion, long-range scene dynamics, and temporally consistent semantic-spatial reasoning. Additional discussion is provided in the appendix.

\section{Acknowledgment}
The authors (MC,ES) acknowledge the partial support of HDR Institute: Accelerated AI Algorithms for Data-Driven Discovery (A3D3) National Science Foundation grant PHY-2117997, National Science Foundation grant EFRI-BRAID-2223495 and National Science Foundation grant PFI-TT-2414896. The authors also acknowledge the partial support by the Departments of Electrical Computer Engineering and Applied Mathematics. The authors are thankful to the eScience Center at the University of Washington.

\clearpage

\bibliography{reference}
\bibliographystyle{unsrt}


\newpage
\appendix

\section{Appendix Overview}
In this appendix, we provide supplementary details and analyses that complement the main paper.
\begin{itemize}

    \item \textbf{Dataset Curation} (Sec.~\ref{app:data}): We describe the data sources, annotation pipeline, verification process, balancing strategy, and benchmark statistics.

    \item \textbf{Implementation Details} (Sec.~\ref{app:impl}): We provide architectural, training, and inference-time matching details for SceneBind.

    \item \textbf{Baseline Fine-tuning} (Sec.~\ref{app:baseline}): We describe how retrieval baselines are trained and evaluated under the same data and benchmark settings.

    \item \textbf{Zero-shot Audio-Visual Localization Details} (Sec.~\ref{app:avloc}): We detail the zero-shot audio-visual localization procedure and evaluation protocol.

  \item \textbf{More Qualitative Results Analysis} (Sec.~\ref{app:qual}): We provide additional comparisons and analyses of SceneBind's semantic-spatial retrieval and grounding behavior.
        
    \item \textbf{Limitations and Future Works} (Sec.~\ref{app:limitations}): We discuss current limitations and future research directions.

    \item \textbf{Broader Impacts} (Sec.~\ref{app:broader_impacts}): We discuss potential societal impacts and safeguards for SceneBind.
\end{itemize}

We also provide an interactive qualitative viewer in our \href{https://scenebind.github.io/}{\texttt{webpage}} for cross modal retrieval and object grounding. It covers all six scene retrieval directions and displays query and retrieved samples with images, binaural audio, and text annotations. We recommend spatial audio playback for audio examples. Details and analysis are provided in Sec.~\ref{app:qual}.

\section{Dataset Curation}
\label{app:data}
SceneBind is trained and evaluated on in-the-wild audio-visual data with aligned semantic and spatial annotations. To support semantic-spatial learning, we curate data from multiple real-world sources and convert them into a unified format containing paired audio-visual samples, object-level text clauses, and spatial labels. This section provides details of the dataset curation process including the data sources, processing pipeline and annotation procedure used for training and evaluation.

\subsection{Data Source}
\textbf{Binaural Dataset.}
We construct the \emph{Binaural} training corpus from in-the-wild binaural-audio videos sourced from publicly available platforms. Each video is cropped to samples where each sample contains a 2-second binaural clip, a representative image frame, and event annotations with semantic and spatial descriptions.
The corpus is designed to support omni-modal spatial learning from in-the-wild scenes.

\textbf{Sphere360 Dataset.}
We additionally construct the \emph{Sphere360} subset from YouTube
$360^{\circ}$ videos with first-order ambisonic (FOA)
audio~\cite{Liu2025OmniAudioGS}. Its $360^{\circ}$ coverage and FOA audio enable
rendering each scene from arbitrary listener orientations, producing
the spatially augmented candidate pool for each query sample.  In our experimental setting, Sphere360 serves exclusively as a zero-shot evaluation benchmark, where candidates share the same
semantics but differ in spatial configuration, testing the model's
ability to discriminate fine-grained spatial structure under semantic
ambiguity. 

\begin{figure}[!htbp]
\centering
\includegraphics[width=1.0\linewidth]{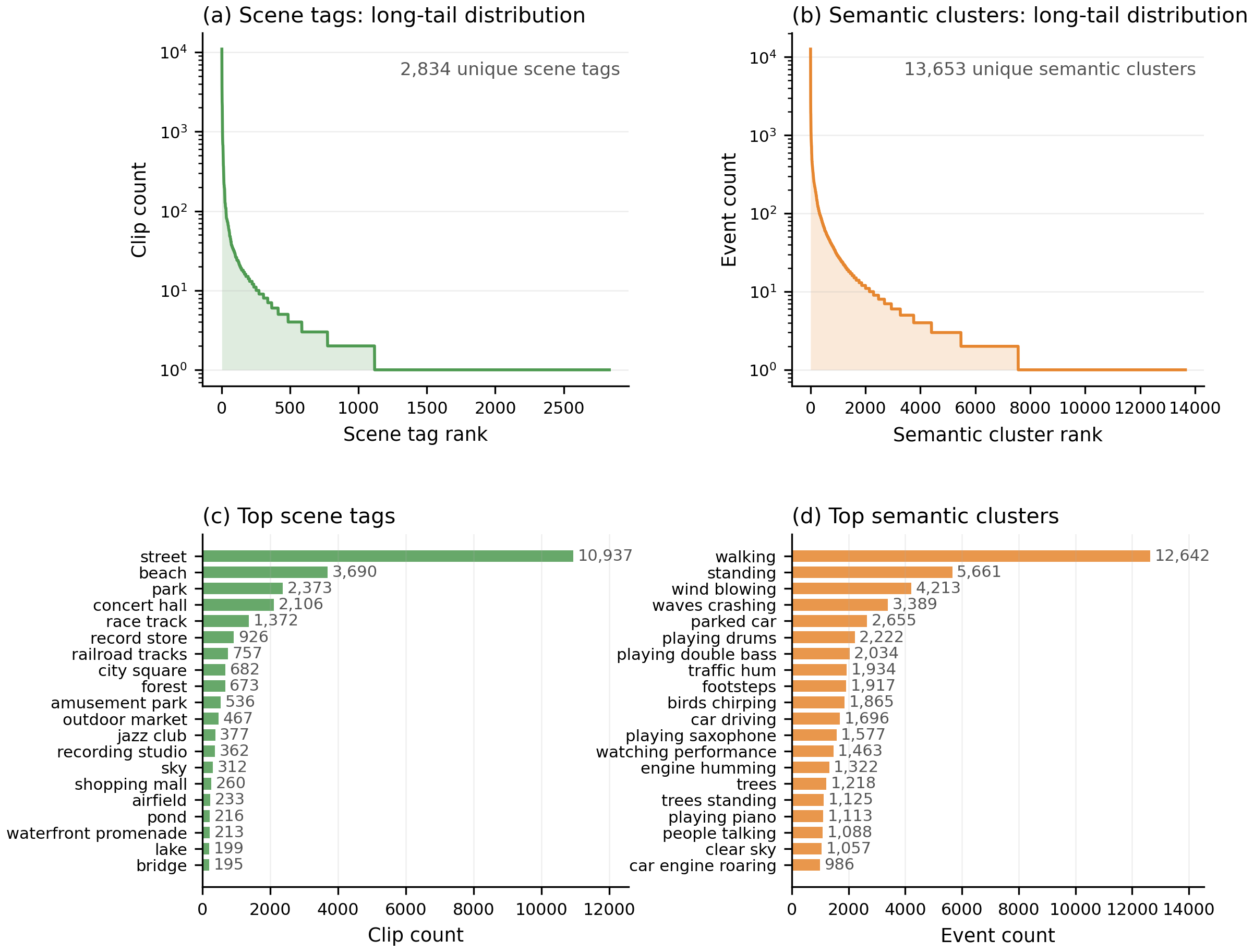}
\caption{Long-tail distributions of scene tags and semantic event
clusters in the Binaural dataset. The top row shows
rank-frequency curves for scene tags and semantic clusters on a
log-scaled count axis. The bottom row reports the 20 most frequent scene tags and semantic clusters. The scene tags span urban, transportation, performance, and outdoor environments, while the semantic clusters cover music, human motion, vehicles, and ambient events. This distribution shows that the curated data contains diverse real-world scenes and a broad range of sound-producing objects and actions.}
\label{fig:tag-longtail}
\end{figure}

\FloatBarrier

\subsection{Pipeline}
\label{app:pipeline}

\begin{figure}[!htbp]
\centering
\includegraphics[width=1.0\linewidth]{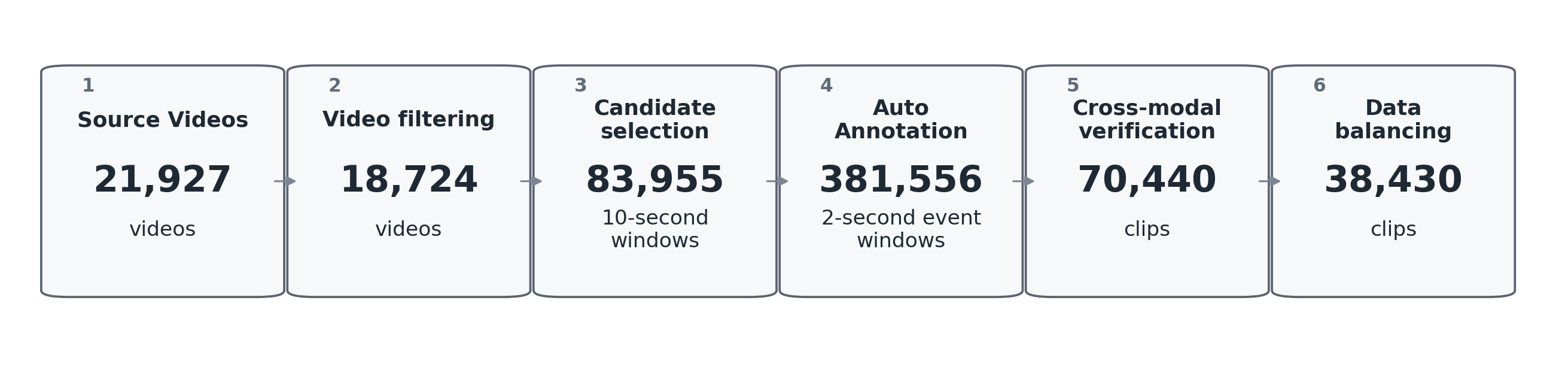}
\caption{Binaural data construction funnel. The pipeline filters source videos, selects diverse candidate windows, annotates short event segments, verifies audio-visual grounding, and applies data balancing to produce the released training set.}
\label{fig:binaural-pipeline}
\end{figure}

We construct the binaural training data with a multi-stage pipeline designed to identify audio-visual events with plausible spatial annotations from source binaural video corpus. The pipeline starts from source in-the-wild videos, filters for usable binaural audio and non-360 content, selects diverse 10-second candidate windows, annotates 2-second event segments, verifies audio-visual grounding with independent encoders, and finally applies data balancing and human review. Figure~\ref{fig:binaural-pipeline} summarizes the resulting data funnel.

Specifically, starting from 21{,}927 source videos, the pipeline retains 18{,}724 after video-level filtering, selects 83{,}955 candidate 10-second windows from 8{,}015 videos via per-clip scoring and Bayesian Gaussian Mixture Models (BayesGMM) diversity selection, annotates 381{,}556 candidate 2-second event windows, verifies 70{,}440 clips with at least one audio-visual event, and produces 56{,}947 clips after spatial balancing.

\textbf{Candidate Construction.}
For the binaural corpus, we first fetch video metadata and retain videos that satisfy deterministic quality checks: landscape orientation, resolution at least 1920 pixels in width or 1080 pixels in height, valid binaural-audio status, and non-360 projection. Each retained video is then divided into short candidate windows in 2 seconds and scored using three complementary signals from pretrained audio-visual models and binaural audio statistics. First, VGGish~\cite{hershey2017cnn} features measure acoustic informativeness and are used to filter out uninformative audio clusters. Second, ImageBind~\cite{girdhar2023imagebind} measures audio-visual alignment, retaining clips whose audio and visual embeddings are sufficiently similar. Third, binaural-cue strength is estimated from frequency-weighted inter-channel level differences and inter-channel coherence. We keep candidates that pass the acoustic cluster filter, have strong audio-visual alignment, and contain reliable binaural cues. Accepted short clips are then grouped into 10-second annotation windows, preserving temporal context for subsequent event annotation and verification.

Candidate construction for Sphere360 is deterministic. We identify each $360^\circ$ clip as either Equirectangular or Equi-Angular Cubemap (EAC) from its metadata, and render four perspective views at $90^\circ$ azimuth intervals with zero pitch and a $90^\circ$ field of view. For each view, we generate the corresponding binaural audio by rotating the first-order ambisonic signal to the view direction and decoding it with the MIT KEMAR HRTF~\cite{Gardner1994HRTFMO} to obtain stereo audio. This produces aligned audio-visual samples with known relative viewing directions.

\textbf{Annotation.}
For the binaural video candidates, we use Gemini~\cite{comanici2025gemini} to generate initial event proposals from each 10-second video window sampled at 2 fps. Although Gemini does not directly perceive binaural spatial cues, it provides strong visual recognition and temporal audio-visual event understanding. We guide it to identify audible events, associate them with visible entities when possible, and estimate the corresponding camera-coordinate spatial labels from the visual evidence. The prompt asks for atomic 1--2 second events, assigns each event to one of three modality types (\texttt{audio\_visual}, \texttt{audio\_only}, or \texttt{visual\_only}), and returns schema-constrained JSON with azimuth, elevation, and distance annotations. This produces plausible spatial annotations for each short event window, which are later filtered by cross-modal verification and spatial balancing. Specifically, we define spatial labels in camera coordinates. Azimuth uses seven coarse directional regions: hard left $[-90^\circ,-60^\circ)$, left $[-60^\circ,-30^\circ)$, slight left $[-30^\circ,-15^\circ)$, front $[-15^\circ,15^\circ]$, slight right $(15^\circ,30^\circ]$, right $(30^\circ,60^\circ]$, and hard right $(60^\circ,90^\circ]$. Elevation uses five vertical regions: high above $[30^\circ,90^\circ]$, above $[10^\circ,30^\circ)$, level $[-10^\circ,10^\circ]$, below $[-30^\circ,-10^\circ)$, and down below $[-90^\circ,-30^\circ)$. Distance is annotated with four ranges: touching/very close $[0,1.5)$ m, close $[1.5,5)$ m, medium $[5,10)$ m, and far $[10,\infty)$ m.  We run Gemini deterministically with temperature $0$. The full prompt for the binaural subset is shown in Fig.~\ref{fig:gemini-prompt}.

For Sphere360, we use a three-round protocol to ensure that annotation accounts for multiple possible viewing directions. 1) In Round 1, Gemini annotates each of the four cardinal views independently with a consistent prompt as shown in Fig.~\ref{fig:gemini-prompt}, and identifies the primary view with the strongest audio-visual correspondence. 2) In Round 2, each candidate event is revisited using five auxiliary neighbor views centered around the primary view, with yaw offsets $\{0^\circ,\pm30^\circ,\pm60^\circ\}$. We input the Round-1 annotations with the neighbor views to Gemini to confirm or revise each event across these viewpoints. 3) In Round 3, we merge cross-view annotations for the same event, retaining the entry whose view places the event closest to the frontal direction, i.e., with the smallest absolute azimuth. The merged event can then be propagated to the rendered views by updating its relative azimuth according to each view direction. This produces augmented audio-visual samples with consistent semantic labels and view-dependent spatial annotations across the 360-degree views.

\FloatBarrier

\begin{figure}[!htbp]
\centering
\fbox{\begin{minipage}{0.94\linewidth}
\footnotesize\ttfamily
Analyze the provided 10-second clip using both AUDIO and
VISUAL cues.\\
Your goal is to identify audio-visual events where visible
objects produce clearly synchronized sound, and group them
by the physical object responsible.\\
Then, to provide complete scene context, enumeratively
identify visual-only and audio-only objects present
\textbf{at the same timestamps of audio\_visual events}.

\medskip
\textbf{1. Event Modality Constraints}\\
- \textbf{audio\_only}: if you can hear the object, but it
is not in the field of view.\\
- \textbf{visual\_only}: if visible, but it is silent or not
synchronized with any sound.\\
- \textbf{audio\_visual}: if visible AND the sound is
clearly synchronized with its motion AND there is strong
audio-visual correspondence.\ \textbf{High precision required.}

\medskip
\textbf{2. Annotation Rules (Strict)}\\
- \textbf{Timestamps:} Must be relative to clip start (0s
to 10s).\\
- \textbf{Duration Constraint:} Events must be short atomic
instances. Duration must be EXACTLY 1 or 2 seconds.\\
- \textbf{Correct Format}: \mbox{[1,3)}, or \mbox{[2,3)}
relative to clip start; \textbf{Incorrect Format}:
\mbox{[00:01:21-00:01:23)} absolute video timestamps.\\
- \textbf{Split Rule}: If an event lasts longer than 2
seconds, split into multiple events; for each split,
consider the event annotation, angle and distance without
any bias from previous or subsequent splits.\\
- \textbf{For audio\_visual event}: When in doubt, leave it
out. False positives are much worse than false negatives.\\
- \textbf{Valid Duration}: When there is at least one valid
audio\_visual event during that 1 or 2 seconds; only
annotate an audio\_only or visual\_only event if the
duration is valid.\\
- \textbf{For visual\_only and audio\_only event}: Be as
enumerative as you can in the valid duration.

\medskip
\textbf{3. Spatial Definitions (Camera Coordinates)}\\
\textbf{Azimuth (Horizontal):} Center is 0 deg; Negative is
left; Positive is right.\\
- Hard Left: \mbox{[-90, -60)}\\
- Left: \mbox{[-60, -30)}\\
- Slight Left: \mbox{[-30, -15)}\\
- Front: \mbox{[-15, 15]}\\
- Slight Right: \mbox{(15, 30]}\\
- Right: \mbox{(30, 60]}\\
- Hard Right: \mbox{(60, 90]}

\medskip
\textbf{Elevation (Vertical):} Level is 0 deg; Negative is
Down; Positive is Up.\\
- High Above: \mbox{[30, 90]}\\
- Above: \mbox{[10, 30)}\\
- Level: \mbox{[-10, 10]}\\
- Below: \mbox{[-30, -10)}\\
- Down Below: \mbox{[-90, -30)}

\medskip
\textbf{4. Field Descriptions}\\
- \textbf{semantic\_tag:} A concise semantic tag for the
event, e.g., `car honking', `playing piano'.\\
- \textbf{semantic\_anno (6 to 8 words):} Describe WHAT the
object is doing/being.\ \textbf{Strictly Forbidden:}
Directional words.\\
- \textbf{spatial\_anno (about 8 words):} Describe WHERE
the object is.\ \textbf{Strictly Forbidden:} Naming the
object, complex action.\\
- \textbf{combined\_anno (about 16 words):} Synthesize both
semantic and spatial details naturally.\\
- \textbf{reason}: Provide the reasoning behind your
estimations for timestamp, semantic\_tag, direction,
height, and distance based on audio and visual cues.
\end{minipage}}
\caption{Audio--Visual event and spatial annotation prompt for binaural videos via Gemini.}
\label{fig:gemini-prompt}
\end{figure}

\FloatBarrier

\textbf{Verification and Balancing.}
\label{app:filtering}
Gemini provides useful event proposals, but its audio-visual associations and spatial estimates can be noisy. To mitigate hallucination, we add a further verification stage to retain short windows whose semantic and spatial annotations are supported by both the visual frame and the audio segment. This step filters hallucinated events, downgrades partially supported events to the appropriate modality type, and balances the spatial distribution of the final data.

\begin{table}[!htb]
\centering
\caption{Event disposition after independent ImageBind (IB) and CLAP verification.}
\label{tab:filter-disposition}
\begin{tabular}{lll}
\toprule
Modality      & Threshold pass & Result            \\
\midrule
{audio-visual} & ImageBind $\And$ CLAP          & keep AV           \\
{audio-visual} & ImageBind        & downgrade to VO   \\
{audio-visual} & CLAP      & downgrade to AO   \\
{audio-visual} & neither        & drop              \\
{audio-only}   & CLAP           & keep AO           \\
{audio-only}   & fail           & drop              \\
{visual-only}  & ImageBind            & keep VO           \\
{visual-only}  & fail           & drop              \\
\bottomrule
\end{tabular}
\end{table}

Each annotated event is verified with pretrained foundation models that are independent of Gemini. Specifically, we compute an ImageBind frame-text score $s^{\mathrm{IB}}_{v\to t}$ and a CLAP~\cite{laion-clap,m2d-clap} audio-text score $s^{\mathrm{CLAP}}_{a\to t}$. An audio-visual event is retained as audio-visual only when both checks pass ($s^{\mathrm{IB}}_{v\to t}\ge 0.10$ and $s^{\mathrm{CLAP}}_{a\to t}\ge 0.05$). If only the visual check passes, the event is downgraded to visual-only; if only the audio check passes, it is downgraded to audio-only; if neither passes, it is discarded. We summarize the verification rules with respect to different original annotated modality in Tab.~\ref{tab:filter-disposition}.

\begin{figure}[!htb]
\centering
\includegraphics[width=0.75\linewidth]{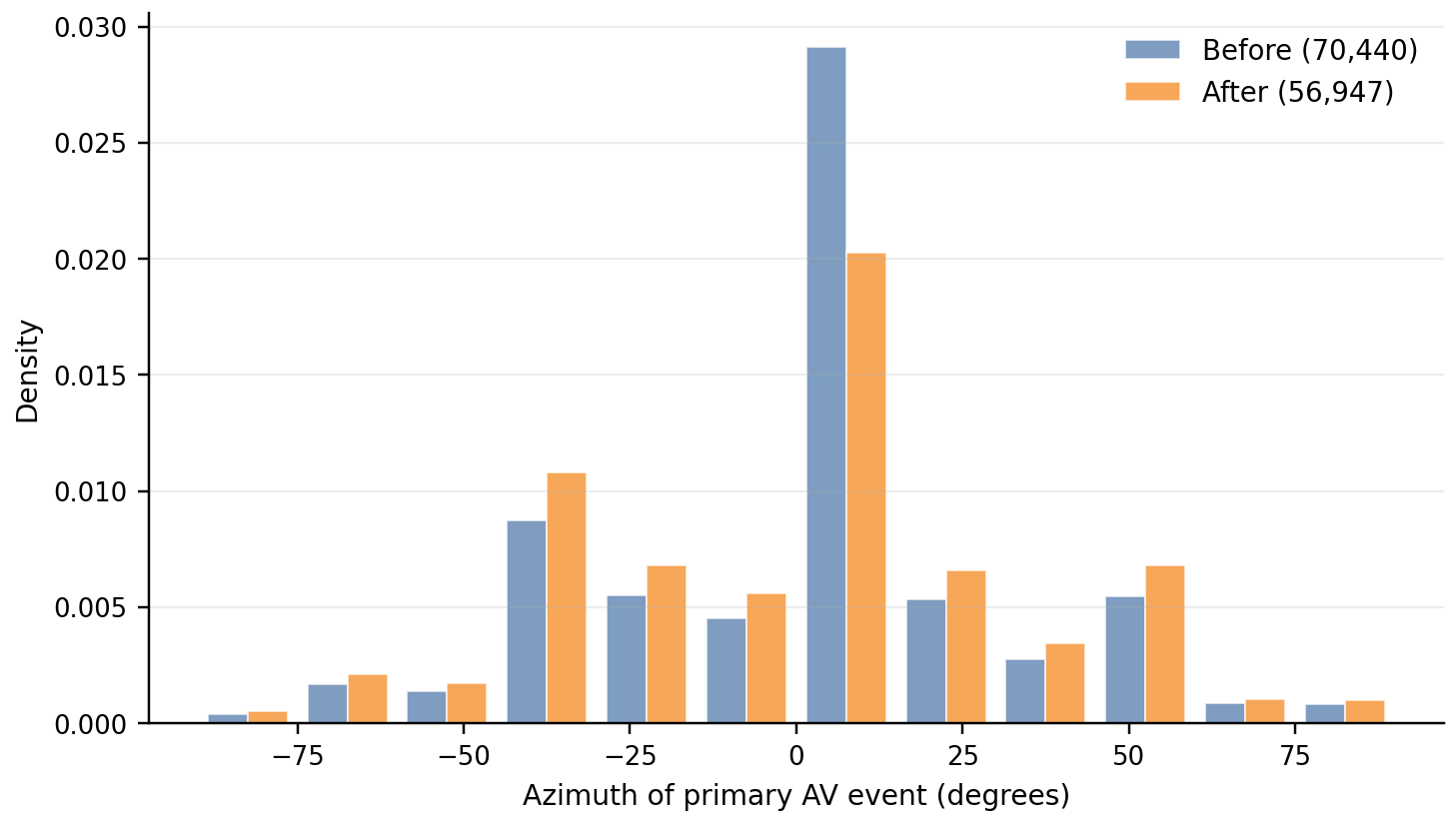}
\caption{Azimuth distribution of the primary audio--visual event per
clip, before ($70{,}440$ clips) and after ($56{,}947$ clips) spatial
balancing. Spatial balancing down-samples the
$\text{azimuth}=0^{\circ}$ bin at a 50\% keep rate, producing a more uniform directional
distribution. }
\label{fig:azimuth-balance}
\end{figure}

After verification, we select the best 2-second event window with the strongest verified event evidence from each 10-second candidate, then choose the representative frame within that window by maximizing ImageBind alignment with the verified audio-visual event labels. We define the selected 2-second audio and its paired image with event and spatial annotations as a clip.
It produces 70{,}440 verified clips with at least one supported audio-visual event. 
For Binaural dataset, to reduce front-facing bias, we down-sample clips whose primary audio-visual event is centered at $\text{azimuth}=0^\circ$ with a 50\% keep rate. This spatial balancing produces the released 56{,}947-clip binaural training set, yields a more uniform directional distribution as shown in Figure~\ref{fig:azimuth-balance}, and discourages models from relying on a front-center shortcut. Sphere360 does not use the spatial balancing step because its $360^\circ$ multi-view augmentation already covers all directions.

We then create train, validation, and test splits at the video level with an 80\%/5\%/15\% ratio, stratifying by channel and semantic cluster to reduce train-test leakage. Specifically for training set, we further balance the training clips to reduce redundancy across scene environment tags, event tags, and YouTube channels, improving diversity. After this diversity balancing, the final training set contains 38{,}430 binaural clips.

\subsection{Human Verification and Interface}
\label{app:human-verification}
We add a human review pass for the binaural benchmark and the zero-shot Sphere360 benchmark. Four reviewers inspect candidate test clips and remove examples with ambiguous content, weak audio-visual correspondence, inaccurate spatial labels, or near-duplicate scenes. When possible, reviewers correct minor spatial-label errors. This final step improves benchmark reliability by retaining clips with meaningful events and reliable semantic-spatial annotations.

\begin{figure}[!htb]
\centering
\includegraphics[width=0.95\linewidth]{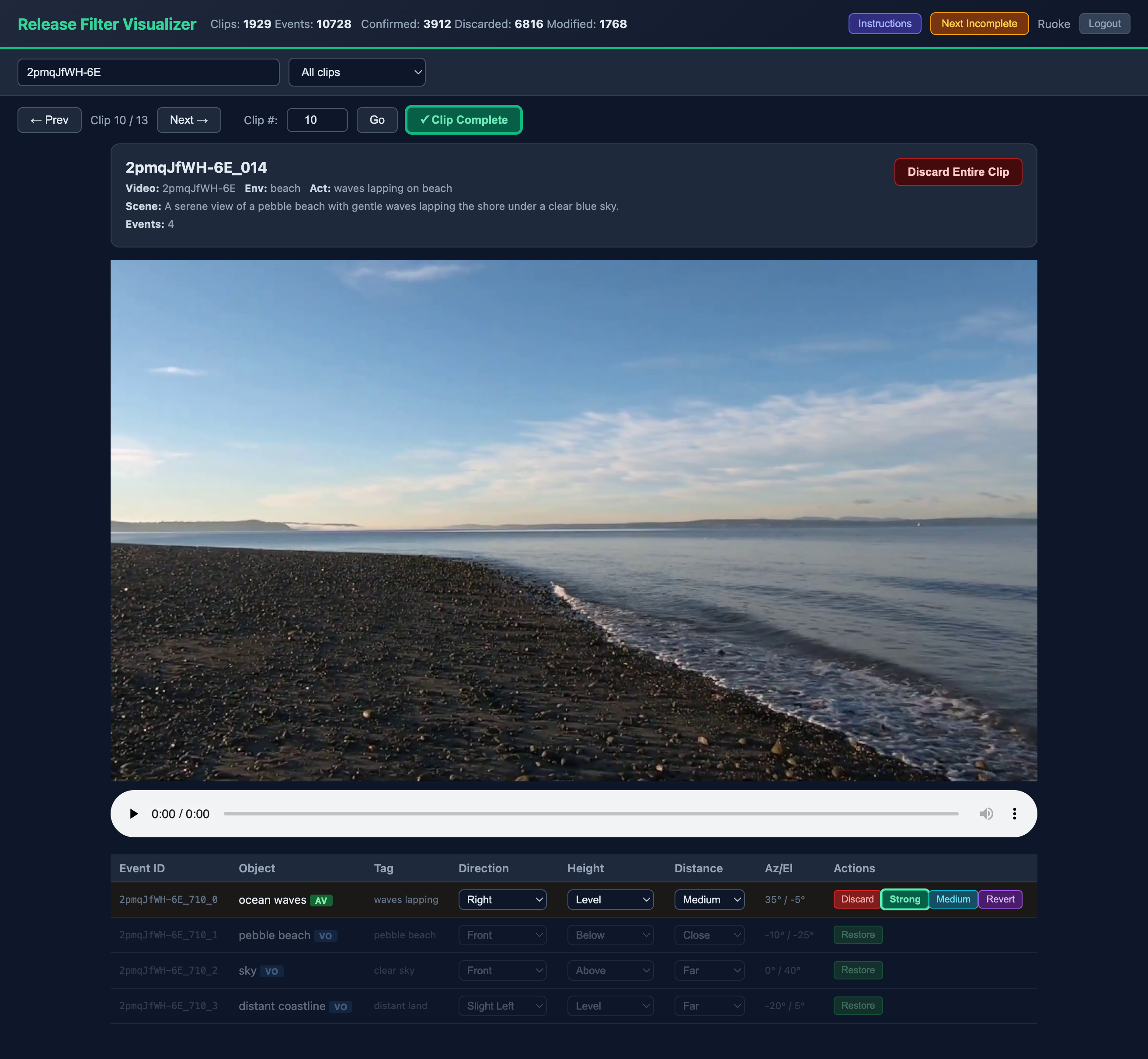}
\caption{Human verification interface for reviewing candidate benchmark clips. It allows reviewers to inspect the representative image, binaural audio, scene description, object clauses, and spatial labels, then discard invalid samples or correct annotations when needed.}
\label{fig:human-evaluation-ui}
\end{figure}

\textbf{Review Interface.}
We build a review interface for clip-level inspection (Fig.~\ref{fig:human-evaluation-ui}). The interface presents the representative image, binaural audio, scene description, object clauses, and associated spatial labels. Reviewers use it to verify audio-visual correspondence, label sample quality, discard invalid examples, and correct spatial annotations when needed. Each reviewer works independently, and the reviewed labels are synchronized in the backend and merged into the final benchmark annotations.

\textbf{Review Guidelines.}
Reviewers first decide whether a candidate clip should be kept. A clip is discarded if it fails any of three criteria: 1) weak relevance, where the audio, visual evidence, and semantic description do not describe the same event; 2) object or event ambiguity, where the annotation cannot be resolved to a clear, unique, and non-trivial entity in the clip; or 3) low sample quality, such as unclear imagery, overly noisy audio, or the absence of meaningful events.

For clips that pass this clip-level check, reviewers then inspect each annotated object. If the semantic annotation is correct and the object has meaningful spatial attributes, reviewers verify and, when necessary, correct its azimuth, elevation, and distance labels. If the object is not semantically valid, spatially meaningful, or uniquely identifiable, it should be removed from the clip.

We ask reviewers to assign a quality label for each object of the retained objects. \emph{Strong} indicates that the audio, visual evidence, semantic and spatial annotation clearly agree and describe a meaningful object or event. \emph{Medium} indicates that the annotation is mostly consistent but contains minor ambiguity or uncertainty.

Finally, reviewers remove near-duplicate clips. When multiple clips show nearly identical scene instances in image, audio, or text, we retain only one clip, prioritizing the sample with clearer events, higher object quality, and greater semantic-spatial diversity.

\subsection{Final Benchmark Details}
After human verification, we compile two evaluation benchmarks from the verified clips: the binaural benchmark and the Sphere360 benchmark. This section summarizes the final split statistics and query-pool construction used for evaluation.

\textbf{Binaural Scene Retrieval.}
The binaural benchmark contains 1{,}066 clips in total. For scene-level retrieval, the pool size can differ by modality because some clips do not contain valid objects for every modality. The audio-visual pool contains 1{,}040 paired samples, while the audio-text and visual-text retrieval pools contain 942 and 1{,}046 clips, respectively.

For general audio-visual retrieval, all paired audio-visual samples are used as queries. For text-based scene retrieval, we focus on non-trivial foreground queries: audio-text queries require more than one audio-relevant object, i.e., audio-visual plus audio-only objects, and visual-text queries require more than one visual-relevant object, i.e., audio-visual plus visual-only objects. This filtering avoids degenerate single-object scenes where the text query does not test object-level disambiguation. Consequently, the resulting benchmark contains 1040 audio-visual scene queries, 325 audio-text scene queries, and 757 visual-text scene queries.

\textbf{Binaural Spatial Retrieval.}
Binaural spatial retrieval uses object-level text queries. Starting from the audio-relevant and visual-relevant object annotations, we retain a query only when its semantic cluster forms a valid hard pool: the pool must contain at least one sample with matching quantized azimuth, elevation, and distance labels, and its size must fall within the predefined range (e.g., $[5,20]$). This yields 745 audio-relevant queries and 782 visual-relevant queries, for 1{,}527 total queries.

Spatial mAP ranks candidates from the full retrieval pool as the scene retrieval task using the joint semantic-spatial object score, and treats samples with matching quantized spatial labels as positives. Spatial-hard R@1 evaluates the same queries in a harder setting: each query is ranked only within its same-cluster hard pool, which contains at most one sample per unique spatial bin. The retained hard pools have mean size 11.75 and median size 10 overall.

\textbf{Binaural Object Grounding.}
Object grounding evaluates the model's predicted object-centric slots against all modality-relevant annotations in the 1{,}066 binaural clips. The benchmark includes 1{,}386 audio-relevant targets from audio-visual and audio-only objects, and 2{,}682 visual-relevant targets from audio-visual and visual-only objects.

\textbf{Sphere360 Retrieval.}
The Sphere360 hard benchmark evaluates spatial disambiguation when objects share similar semantics but appear from different viewing directions. The split contains 97 clips and 4{,}277 annotated objects, including 553 audio-visual objects. Query objects are restricted to frontal examples with azimuth in $[-90^\circ,90^\circ]$, while candidate pools are drawn from augmented 360-degree views. The per-query candidate pool has mean size 45.4 and median size 47.

\section{SceneBind Implementation Details}
\label{app:impl}
\subsection{SceneBind Encoding Model}
\textbf{Visual and Text Encoder.}
We use a frozen SigLIP2~\cite{Tschannen2025SigLIP2M} visual backbone as the visual feature extractor. For each frame, the backbone produces 256 patch tokens in a 1152-dimensional embedding space, which serve as the visual context tokens. During training, we apply horizontal flip augmentation by using the corresponding flipped visual features and mirroring the azimuth label.

For text encoding, we use the frozen text tower of the same SigLIP2 model. We encode two types of text supervision. First, the scene description is encoded as a single global 1152-dimensional text embedding, which supervises global semantics alignment. Second, each object clause is encoded independently as a 1152-dimensional semantic embedding, which matches with audio and visual object slots. Each object clause is also associated with spatial labels $\mathbf{r}_k=(\theta_k,\phi_k,d_k)$, represented as one-hot targets over the discretized azimuth, elevation, and distance bins.

\textbf{Audio Semantics Encoder.}
The binaural waveform is loaded as a two-channel signal and padded or cropped to the fixed audio window. For semantic audio encoding, we average the two channels to obtain a mono waveform and resample it to 16 kHz. A frozen M2D-CLAP audio encoder~\cite{m2d-clap} extracts a sequence of 768-dimensional audio patch tokens. We then use a residual MLP alignment module to project these tokens into the 1152-dimensional SigLIP2 embedding space. The alignment module contains five residual blocks with hidden width 1024 and LayerNorm/GELU nonlinearities. It is pretrained to align M2D-CLAP audio features with SigLIP2 text features on the Binaural training captions, and both the M2D backbone and the alignment module are frozen during SceneBind training. After alignment, the semantic audio representation has shape $310 \times 1152$ for each 10-second audio window. These 310 tokens correspond to a $5 \times 62$ spectro-temporal patch grid over the M2D log-mel representation, with 5 frequency patches and 62 time patches.

\textbf{Audio Spatial Encoder.}
For spatial audio encoding, we use the original binaural waveform. The binaural signal is resampled to 16 kHz and converted into a four-channel time-frequency representation consisting of left/right magnitude and phase features from an STFT with FFT size 1024 and hop size 512. A convolutional frontend processes this representation with three convolutional blocks, Each block uses a 3-by-3 convolution with 64 output channels, ReLU, batch normalization, and max pooling on the frequency axis. SceneBind uses the resulting convolutional feature map as the spatial patch representation. Since the frontend operates on an STFT frequency grid that is not aligned with the M2D patch grid, we average over the remaining frequency dimension to obtain a temporal sequence of 64-dimensional binaural features with frequency-aggregated spatial cues. Each feature is projected to the shared 1152-dimensional space and linearly interpolated to length 310, forming spatial audio tokens that match the flattened M2D semantic token sequence.

\textbf{Semantic Decoder.}
For each modality, SceneBind uses a query-based semantic decoder. We initialize one learnable global query and $K=50$ learnable object queries. A one-layer Transformer decoder cross-attends these learnable queries to the modality-specific semantic context tokens. The global-query output is used as the global semantic embedding, while the object-query outputs define semantic embeddings for predicted object slots. Each object slot is additionally passed through a confidence head, producing a score in $[0,1]$ that indicates whether the slot corresponds to an active object.

Let $H^m$ denote the semantic context tokens for modality $m$, and let $Q=[q_{\mathrm{global}},q_1,\ldots,q_K]$ denote the learnable decoder queries. The semantic decoder computes
\[
    [\mathbf{s}^m_{\mathrm{global}}, \mathbf{s}^m_1,\ldots,\mathbf{s}^m_K]
    =
    \operatorname{Dec}_{\mathrm{sem}}(Q; H^m),
\]
where $\mathbf{s}^m_{\mathrm{global}}$ is the global embedding and $\mathbf{s}^m_i$ is the semantic embedding of object slot $i$.

\textbf{Spatial Decoder.}
The spatial decoder predicts spatial attributes for each object slot by conditioning its semantic embedding on modality-specific spatial evidence. Its output is passed through shared spatial projection heads and classifiers for azimuth, elevation, and distance. The classifiers produce distributions $\boldsymbol{\pi}^{m,\theta}_i$, $\boldsymbol{\pi}^{m,\phi}_i$, and $\boldsymbol{\pi}^{m,d}_i$ over 7, 5, and 4 bins (defined in Sec.~\ref{app:pipeline}), respectively; their argmax bins form $\hat{\mathbf{r}}^m_i=(\hat{\theta}^m_i,\hat{\phi}^m_i,\hat{d}^m_i)$ (defined in Sec.~\ref{app:pipeline}).

For the visual branch, semantic object slots serve as decoder queries, while visual context tokens from the visual encoder serve as memory:
\[
    h^v_i = \operatorname{Dec}_{\mathrm{spa}}(\mathbf{s}^v_i; H^v),
    \qquad
    \hat{\mathbf{r}}^v_i = g_{\mathrm{spa}}(h^v_i),
\]
where $h^v_i$ is the decoded spatial feature and $g_{\mathrm{spa}}$ denotes the shared spatial heads.

For the audio branch, semantic object slots again serve as decoder queries. The decoder memory is the fused audio context, obtained by adding semantic audio tokens and length-aligned spatial audio tokens:
\[
    H^a_{\mathrm{fused}}
    =
    H^a_{\mathrm{sem}}
    +
    \widetilde{H}^a_{\mathrm{spatial}},
    \qquad
    h^a_i
    =
    \operatorname{Dec}_{\mathrm{spa}}(\mathbf{s}^a_i; H^a_{\mathrm{fused}}),
    \qquad
    \hat{\mathbf{r}}^a_i = g_{\mathrm{spa}}(h^a_i).
\]
Thus, the semantic slots first form object-level hypotheses, and the spatial decoder then retrieves visual or spatial audio evidence for each hypothesized object. This enables SceneBind to handle variable numbers of objects or events within a scene.

\subsection{Training Objective Details}
We train SceneBind with a combination of global-level semantic alignment, object-centric semantic-spatial grounding, and cross-modal semantic-spatial contrastive losses.

\textbf{Global Alignment.}
Scene-level alignment is trained with a symmetric InfoNCE~\cite{infonce} loss between paired modalities. Let $S^m_{\mathrm{global}}\in\mathbb{R}^{B\times d}$ denote the batch-stacked normalized global embeddings for modality $m$. For paired modalities $m$ and $n$, we compute
\[
    \mathcal{L}_{\mathrm{global}}^{m,n}
    =
    \frac{1}{2}
    \left[
    \operatorname{CE}\left(
    \frac{S^m_{\mathrm{global}}(S^n_{\mathrm{global}})^\top}{\tau}, I
    \right)
    +
    \operatorname{CE}\left(
    \frac{S^n_{\mathrm{global}}(S^m_{\mathrm{global}})^\top}{\tau}, I
    \right)
    \right],
    \qquad \tau=0.07 .
\]
This loss is applied whenever paired sample-level supervision is available. For text, global alignment uses the scene-description embedding.

\textbf{Object Grounding Supervision.}
After assigning ground-truth object clauses to predicted object slots by bipartite matching, we supervise each matched slot with the semantic, confidence, and spatial losses defined in Sec.~\ref{sec:training}. The semantic loss is a cosine-distance loss between the predicted slot embedding and the matched object-clause embedding. The confidence loss is binary cross-entropy, where matched slots are treated as positives and unmatched slots as negatives. Spatial attributes are supervised with Gaussian-smoothed cross-entropy over discretized azimuth, elevation, and distance bins. We use circular smoothing with $\sigma=1.0$ for azimuth and non-circular smoothing with $\sigma=0.5$ for elevation and distance.

\textbf{Intra-Scene Contrastive Loss $\mathcal{L}_{inst}$.}
For an audio-visual pair, slots from the two modalities are first matched to the same ground-truth object clauses.  Co-matched slot pairs are treated as positives within the scene.  If multiple objects are present, we apply a symmetric contrastive loss over the co-matched audio and visual slots, so that each object is closest to its counterpart in the other modality.
The same principle is used for audio-text and visual-text grounding: the modality slot matched to an object is treated as a positive pair with the corresponding object-clause text embedding, while other clauses in the same scene serve as negatives. If only one object is co-matched, the loss reduces to paired cosine alignment.

\textbf{Cross-Scene Contrastive Loss $\mathcal{L}_{\mathrm{batch}}$.}
To improve object-level discriminability across scenes, SceneBind applies batch-wise slot contrastive losses over slots matched to ground-truth object clauses. For audio-visual training, audio and visual slots independently matched to the same ground-truth object form positive pairs. We apply symmetric InfoNCE to both semantic slots and spatial-head embeddings:
\[
\mathcal{L}^{a,v}_{\mathrm{batch}}
=
\mathcal{L}_{\mathrm{NCE}}(\mathbf{s}^a,\mathbf{s}^v)
+
\mathcal{L}_{\mathrm{NCE}}(\mathbf{u}^{a,\theta},\mathbf{u}^{v,\theta})
+
\frac{1}{2}\mathcal{L}_{\mathrm{NCE}}(\mathbf{u}^{a,\phi},\mathbf{u}^{v,\phi})
+
\frac{1}{2}\mathcal{L}_{\mathrm{NCE}}(\mathbf{u}^{a,d},\mathbf{u}^{v,d}),
\]
where $\mathbf{u}^{\theta}$, $\mathbf{u}^{\phi}$, and $\mathbf{u}^{d}$ are normalized spatial-head embeddings before the final classifiers. 

For audio-text and visual-text pairs, the cross-scene contrastive loss is semantic-only because text clauses do not provide learned spatial-head features:
\[
\mathcal{L}^{a,t}_{\mathrm{batch}}
=
\mathcal{L}_{\mathrm{NCE}}(\mathbf{s}^{a},\mathbf{s}^{t}),
\qquad
\mathcal{L}^{v,t}_{\mathrm{batch}}
=
\mathcal{L}_{\mathrm{NCE}}(\mathbf{s}^{v},\mathbf{s}^{t}).
\]

\textbf{Full Loss.}
The final objective is a weighted sum of the losses described above. Table~\ref{tab:loss_weights} summarizes the loss terms and weights used in the main training setting.

\begin{table}[h]
\centering
\small
\caption{Loss weights used in the main SceneBind training setting.}
\begin{tabular}{llc}
\midrule
Category & Loss term & Weight \\
\midrule
Object grounding & Semantic grounding $\mathcal{L}_{\mathrm{sem}}$ & 1.0 \\
Object grounding & Confidence $\mathcal{L}_{\mathrm{conf}}$ & 2.0 \\
Object grounding & Azimuth $\mathcal{L}_{\theta}$ & 1.0 \\
Object grounding & Elevation $\mathcal{L}_{\phi}$ & 0.5 \\
Object grounding & Distance $\mathcal{L}_{d}$ & 0.5 \\
\midrule
Global alignment & Audio-visual $\mathcal{L}^{a,v}_{\mathrm{global}}$ & 2.0 \\
Global alignment & Audio-text $\mathcal{L}^{a,t}_{\mathrm{global}}$ & 1.0 \\
Global alignment & Visual-text $\mathcal{L}^{v,t}_{\mathrm{global}}$ & 1.0 \\
\midrule
Intra-scene contrastive & Audio-visual $\mathcal{L}^{a,v}_{\mathrm{inst}}$ & 1.0 \\
Intra-scene contrastive & Audio-text $\mathcal{L}^{a,t}_{\mathrm{inst}}$ & 1.0 \\
Intra-scene contrastive & Visual-text $\mathcal{L}^{v,t}_{\mathrm{inst}}$ & 0.5 \\
\midrule
Cross-scene contrastive & Audio-visual $\mathcal{L}^{a,v}_{\mathrm{batch}}$ & 2.0 \\
Cross-scene contrastive & Audio-text $\mathcal{L}^{a,t}_{\mathrm{batch}}$ & 0.5 \\
Cross-scene contrastive & Visual-text $\mathcal{L}^{v,t}_{\mathrm{batch}}$ & 0.5 \\
\midrule
\end{tabular}
\label{tab:loss_weights}
\end{table}

\subsection{Training Stages and Data Recipe}
\label{app:train_recipe}
SceneBind is trained in two stages. Stage 1 learns a shared audio-visual-text space from mixed supervision, while Stage 2 specializes the model on Binaural dataset.

\textbf{Stage 1.}
Stage 1 trains for 30 epochs with three data sources: Binaural scene dataset, AudioCaps~\cite{audiocaps}, and MS-COCO~\cite{lin2014microsoft}. Dataset sampling follows
\[
p_i \propto n_i^{0.5} w_i,
\]
where $n_i$ is the dataset size and $w_i$ is the dataset weight. The paired binaural data receive the largest weight because they provide the full spatially grounded audio-visual signal. AudioCaps provides audio-text semantic supervision, and MS-COCO provides visual-text semantic supervision with partial spatial labels, where only azimuth and elevation supervision are used.

\textbf{Stage 2.}
Stage 2 initializes from the Stage 1 model weights, resets the optimizer, and trains for another 30 epochs using only the paired binaural scene dataset.

\textbf{Optimization and Compute.}
Both stages use AdamW with weight decay $0.01$, cosine learning-rate decay, and gradient clipping at norm $1.0$. Stage 1 uses learning rate $10^{-4}$, 200 warmup steps, and global batch size 1024. Stage 2 uses learning rate $5\times10^{-5}$, 100 warmup steps, and global batch size 512. The reported run was trained on one node with 8 NVIDIA H200 GPUs, each with 140GB memory. Using 8 data-loading workers per GPU process, the two-stage run completed in roughly 4 wall-clock hours on this setup.

\begin{table}[h]
\centering
\small
\captionof{table}{Training data recipe and active supervision. $n_i$ is the source dataset size and $w_i$ is the dataset weight.}
\begin{tabular}{lllrrl}
\toprule
Stage & Data source & $n_i$ & $w_i$ & Sample  & Active supervision \\
\midrule
1 & Binaural scenes & 38{,}430 & 2.0 & 54.8\%&
$\mathcal{L}_{\mathrm{global}}$,
$\mathcal{L}_{\mathrm{sem}}$,
$\mathcal{L}_{\mathrm{conf}}$,
$\mathcal{L}_{\theta,\phi,d}$,
$\mathcal{L}_{\mathrm{inst}}$,
$\mathcal{L}_{\mathrm{batch}}$ \\
1 & AudioCaps & 91{,}254 & 0.5 & 21.1\% &
$\mathcal{L}_{\mathrm{global}}$,
$\mathcal{L}_{\mathrm{sem}}$,
$\mathcal{L}_{\mathrm{conf}}$, \\
1 & MS-COCO & 118{,}248 & 0.5 & 24.1\% &
$\mathcal{L}_{\mathrm{global}}$,
$\mathcal{L}_{\mathrm{sem}}$,
$\mathcal{L}_{\mathrm{conf}}$,
$\mathcal{L}_{\theta,\phi}$ \\
\midrule
2 & Binaural scenes & 38{,}430 & 1.0 & 100.0\% &
$\mathcal{L}_{\mathrm{global}}$,
$\mathcal{L}_{\mathrm{sem}}$,
$\mathcal{L}_{\mathrm{conf}}$,
$\mathcal{L}_{\theta,\phi,d}$,
$\mathcal{L}_{\mathrm{inst}}$,
$\mathcal{L}_{\mathrm{batch}}$ \\
\bottomrule
\end{tabular}
\label{tab:training_recipe}
\end{table}

\subsection{SceneBind Matching Details}
\label{app:matching}
At inference time, each sample from modality $m$ is represented by $\mathcal{X}^m$, containing a global semantic embedding and a set of object-centric semantic-spatial slots. We first apply a confidence gate, keeping slots with confidence above $0.05$ as active slots. If no slot passes the gate, we retain the highest-confidence slot. SceneBind Matching then computes two complementary scores: a global score $S_{\mathrm{global}}$ from the cosine similarity between global embeddings, and an object-centric score $S_{\mathrm{obj}}$ from matched semantic-spatial slots of paired scenes. We propose different strategy to use these scores for different tasks such as scene retrieval, spatial retrieval and object grounding.

\textbf{Object Slot Score.}
Given a query scene $q$ and candidate scene $x$, we compare each active query slot $i$ with each active candidate slot $j$ using a semantic-spatial score:
\[
    R_{ij}(q,x)
    =
    \cos(\mathbf{s}^q_i,\mathbf{s}^x_j)
    \cdot
    \rho_{ij},
\]
where $\mathbf{s}^q_i$ and $\mathbf{s}^x_j$ are the semantic embeddings of the two slots. The spatial term is
\[
    \rho_{ij}
    =
    \frac{
    2\langle\boldsymbol{\pi}^{q,\theta}_i,\boldsymbol{\pi}^{x,\theta}_j\rangle
    + \langle\boldsymbol{\pi}^{q,\phi}_i,\boldsymbol{\pi}^{x,\phi}_j\rangle
    + \langle\boldsymbol{\pi}^{q,d}_i,\boldsymbol{\pi}^{x,d}_j\rangle
    }{4}.
\]
Here $\boldsymbol{\pi}^{\theta}$, $\boldsymbol{\pi}^{\phi}$, and $\boldsymbol{\pi}^{d}$ denote the softmax-normalized probability distributions over azimuth, elevation, and distance bins, respectively. The inner products measure distributional agreement along each spatial axis, with azimuth given double weight because horizontal direction is the primary cue in our binaural setting.

\textbf{Best-Match Object Score.}
After computing the pairwise slot scores $R_{ij}(q,x)$ between the query scene $q$ and a candidate scene $x$, each active query slot from $q$ selects the candidate slot from $x$ with the highest confidence-weighted semantic-spatial score:
\[
j_i^\star
=
\arg\max_j
w_{ij}R_{ij}(q,x),
\qquad
w_{ij}=\sqrt{c_i^q c_j^x},
\]
where $c_i^q$ and $c_j^x$ are the confidence scores of the query and candidate slots, and $w_{ij}$ is their geometric-mean confidence. We then aggregate the selected matches into an object-centric sample score:
\[
    S_{\mathrm{obj}}(q,x)
    =
    \frac{
    \sum_i w_{i j_i^\star} R_{i j_i^\star}(q,x)
    }{
    \sum_i w_{i j_i^\star}+\epsilon
    },
    \qquad \epsilon=10^{-8}.
\]
Here the summation is over active query slots, and $j_i^\star$ denotes the selected candidate slot for query slot $i$. Confidence weighting emphasizes reliable slot pairs, while independent best matching supports partial object overlap and variable object counts.

\textbf{Scene Retrieval.}
For scene retrieval, we combine the global embedding score with the object-centric score. Let
$S_{\mathrm{global}}(q,x)=\cos(\mathbf{s}^{q}_{\mathrm{global}},\mathbf{s}^{x}_{\mathrm{global}})$
denote the global similarity between query scene $q$ and candidate scene $x$. We first rank candidates by $S_{\mathrm{global}}$ and select the top 50 candidates for object-level reranking. For each query, we min-max normalize $S_{\mathrm{global}}$ and $S_{\mathrm{obj}}$ over the candidate set, denoted as $\bar{S}_{\mathrm{global}}$ and $\bar{S}_{\mathrm{obj}}$. The final scene similarity is
\[
S_{\mathrm{scene}}(q,x)
=
\bar{S}_{\mathrm{global}}(q,x)
+
\lambda_{\mathrm{obj}}\bar{S}_{\mathrm{obj}}(q,x),
\]
for candidates in the top-50 set, while all remaining candidates keep their global score. We use $\lambda_{\mathrm{obj}}=0.05$ for audio-visual retrieval and $\lambda_{\mathrm{obj}}=0.5$ for text-related retrieval, since text queries benefit more from object-level semantic-spatial matching.

\textbf{Spatial Retrieval.}
Spatial retrieval uses an object clause as the query. The text query provides one-hot spatial distributions $\boldsymbol{\pi}^{t,\theta}$, $\boldsymbol{\pi}^{t,\phi}$, and $\boldsymbol{\pi}^{t,d}$ for azimuth, elevation, and distance. We score a candidate sample $x$ by its best active slot using spatial agreement only:
\[
    S_{\mathrm{spa}}(t,x)
    =
    \max_j
    c^x_j
    \cdot
    \frac{
    2\langle\boldsymbol{\pi}^{t,\theta},\boldsymbol{\pi}^{x,\theta}_j\rangle
    + \langle\boldsymbol{\pi}^{t,\phi},\boldsymbol{\pi}^{x,\phi}_j\rangle
    + \langle\boldsymbol{\pi}^{t,d},\boldsymbol{\pi}^{x,d}_j\rangle
    }{4},
\]
where $c^x_j$ is the confidence of candidate slot $j$, and $\boldsymbol{\pi}^{x,\theta}_j$, $\boldsymbol{\pi}^{x,\phi}_j$, and $\boldsymbol{\pi}^{x,d}_j$ are the softmax-normalized predicted distributions over azimuth, elevation, and distance bins.

\textbf{Object Grounding.}
For object grounding, the query is an object-clause text semantic description encoded as a text embedding $\mathbf{s}^{t}$. We first apply the same confidence gate to the predicted slots of the queried scene, retaining slots with $c_i>0.05$ and falling back to the highest-confidence slot if none pass. Among the retained slots, we select the slot with the largest semantic-confidence score:
\[
    i^\star
    =
    \arg\max_i
    c_i\cos(\mathbf{s}_i,\mathbf{s}^{t}).
\]
If the selected slot has confidence below $0.05$ or semantic similarity below $0.5$, the queried object is treated as missing. Otherwise, we take the argmax of the selected slot's azimuth, elevation, and distance logits as the predicted spatial bins, and compare them with the ground-truth labels.

\section{Baseline Finetune Details}
\label{app:baseline}
We fine-tune all retrieval baselines on the same data sources and splits as SceneBind, following Table~\ref{tab:training_recipe}. This gives all methods access to the same audio-visual samples, scene descriptions, and object-clause annotations. Unlike SceneBind, the baselines produce a single embedding per modality sample and have no spatial grounding heads, so spatial attributes cannot be supervised as separate structured labels. We therefore incorporate the spatial attributes into the object-clause text itself, yielding textual queries that describe both object semantics and spatial location. This setting is used for all baselines in Sec.~\ref{sec:baselines}.

\textbf{Training Objective.}
For paired modalities $m$ and $n$, each baseline is optimized with the same symmetric InfoNCE objective used for global alignment. We apply this loss to audio-visual, audio-text, and visual-text pairs whenever the corresponding supervision is available. Since baselines produce one embedding per modality sample rather than object slots, we use object-clause annotations through a multi-positive clip-to-clause contrastive loss.

Let $\mathbf{s}_b$ be the normalized embedding of sample $b$ in a minibatch of size $B$, and let $\{\mathbf{t}_{\ell}\}_{\ell=1}^{L}$ be all normalized object-clause text embeddings in the minibatch. We denote by $o(\ell)$ the sample that owns clause $\ell$. The multi-positive clause loss is
\[
\mathcal{L}_{\mathrm{clause}}
=
\frac{1}{2}
\left[
-\frac{1}{B}\sum_{b=1}^{B}
\log
\frac{
\sum_{\ell:o(\ell)=b}
\exp(\mathbf{s}_b^\top \mathbf{t}_{\ell}/\tau)
}{
\sum_{\ell=1}^{L}
\exp(\mathbf{s}_b^\top \mathbf{t}_{\ell}/\tau)
}
+
\operatorname{CE}
\left(
\frac{T S^\top}{\tau},
o
\right)
\right],
\qquad \tau=0.07 .
\]
Here $S$ stacks the $B$ sample embeddings, $T$ stacks the $L$ clause embeddings, and $o$ is the clause-to-sample target index. The first term pulls each sample toward all of its own object clauses, while the second term assigns each clause back to its owning sample.

The clause loss treats all object clauses from the same sample as positives for the corresponding sample embedding, and also assigns each clause back to its owning sample. Table~\ref{tab:baseline_loss_weights} summarizes the baseline fine-tuning losses and their weights. Thus, baselines receive both global-level and object-clause text supervision, but do not perform object-centric prediction or spatial decoding.

\begin{table}[t]
\centering
\small
\caption{Fine-tuning losses for global-embedding baselines. Global losses use one paired embedding per sample, while clause losses use all valid object-clause embeddings associated with the sample.}
\begin{tabular}{llc}
\midrule
Loss term & Supervision source & Weight \\
\midrule
$\mathcal{L}^{a,v}_{\mathrm{global}}$ & Paired audio and visual sample embeddings & 2.0 \\
$\mathcal{L}^{a,t}_{\mathrm{global}}$ & Audio embedding and scene-description text embedding & 1.0 \\
$\mathcal{L}^{v,t}_{\mathrm{global}}$ & Visual embedding and scene-description text embedding & 1.0 \\
$\mathcal{L}^{a,t}_{\mathrm{clause}}$ & Audio embedding and object-clause text embeddings & 0.5 \\
$\mathcal{L}^{v,t}_{\mathrm{clause}}$ & Visual embedding and object-clause text embeddings & 0.5 \\
\midrule
\end{tabular}
\label{tab:baseline_loss_weights}
\end{table}

\textbf{ImageBind.}
For ImageBind~\cite{girdhar2023imagebind}, we use the pretrained ImageBind audio and image encoders to obtain audio and visual embeddings. We add lightweight trainable projection heads to map these embeddings into the shared retrieval space. The pretrained backbone is kept frozen in the standard fine-tuning setting, and only the retrieval heads are optimized.

\textbf{M2D+SigLIP2.}
For M2D~\cite{m2d-clap}+SigLIP2~\cite{Tschannen2025SigLIP2M}, audio is encoded with the pretrained M2D-CLAP audio encoder and visual/text features are encoded with SigLIP2. A trainable retrieval head maps the audio representation into the SigLIP2-aligned retrieval space. In the fair comparison setting, the pretrained M2D backbone and the audio-to-SigLIP2 alignment module are frozen, and only the retrieval head is trained.

\textbf{SpatialCLAP+SigLIP2.}
For SpatialCLAP~\cite{spatialclap}+SigLIP2~\cite{Tschannen2025SigLIP2M}, binaural audio is encoded by the pretrained SpatialCLAP audio encoder, while visual and text features are encoded by SigLIP2. We add trainable projection heads to align the audio and visual embeddings in a common retrieval space. The pretrained audio alignment module is frozen in the fair comparison setting.

All baselines follow the same two-stage fine-tuning schedule as SceneBind. Stage 1 trains for 30 epochs on the mixed data recipe with paired Binaural dataset, AudioCaps, and MS-COCO, with a batch size of 1024. Stage 2 initializes from the Stage 1 weights and trains for another 30 epochs on the paired Binaural dataset only, with batch size equals to 512. We use AdamW with weight decay $0.01$, cosine learning-rate decay, warmup, and gradient clipping at norm $1.0$. The learning rate is $10^{-4}$ in Stage 1 and $5\times10^{-5}$ in Stage 2, matching the SceneBind schedule (Table~\ref{tab:training_recipe} and Sec.~\ref{app:train_recipe}). 

Baselines are evaluated with the same retrieval pools, benchmark splits, and sample-level protocol as SceneBind. At inference time, each baseline produces a single embedding per modality sample. To keep the scoring protocol consistent with SceneBind Matching~\ref{app:matching}, we use this embedding both as the global representation and as a one-slot object representation for object-level matching. Since the baselines do not predict spatial attributes, they cannot explicitly compare azimuth, elevation, or distance labels. For spatially conditioned text clause queries, we instead encode the combined semantic-spatial description as one text embedding and match it directly to the sample embedding.

\section{Zero-Shot Audio-Visual Localization Details}
\label{app:avloc}
For the zero-shot audio-visual localization task in Sec.~\ref{sec:avloc}, the input is a stereo audio clip and an image frame. We evaluate SceneBind without finetuning. The audio branch is run once on the stereo audio to produce audio object slots, while the visual branch is run on the image to produce visual object slots and the global semantic query's cross-attention over visual patch tokens. This global cross-attention map provides a coarse visual support map for candidate sounding-object regions.

In this task, the sounding object should match a visual object in semantics. We therefore score audio-visual slot pairs by semantic similarity and slot confidence, and use the best-matched audio slot to weight visual cross-attention maps.
We then construct spatial guide maps for the selected pair on the image plane. Since camera intrinsics are unavailable, we use a normalized camera-coordinate projection: the image center corresponds to azimuth $0^\circ$ and elevation $0^\circ$, horizontal position is determined by azimuth, and vertical position is determined by elevation. Under this projection, the selected visual slot's predicted azimuth and elevation define a center on a $72\times72$ grid, where we render a 2D Gaussian guide map with $\sigma_x=\sigma_y=0.13$.

For the selected audio slot, we build an analogous guide map from its predicted spatial distributions. We temperature-scale the azimuth, elevation, and distance logits with various temperatures (e.g., $0.45$, $0.60$, and $0.75$), respectively, before softmax normalization. The audio guide center is estimated from the predicted azimuth/elevation bin and its immediate neighboring bins. The distance distribution controls the Gaussian width through distance-dependent base scales, with a small confidence-dependent broadening.

The visual and audio guide maps are normalized and fused into an audio-visual support map, indicating where the matched audio-visual slot pair agrees the sounding object should lie in the image. We use this support map to select the relevant region from the visual cross-attention map. The cross-attention map is thresholded at multiple levels to produce connected-component candidates, and each component is scored by its overlap with the support map and the strongest audio-visual overlap seed. We keep the highest-scoring component and suppress the rest. The final localization map therefore preserves the structure of the visual attention map while selecting the region most consistent with the matched audio-visual spatial evidence.

For cIoU evaluation, the final heatmap is normalized and thresholded at $0.4$. If multiple connected components remain, we keep the component with the strongest support from the audio-visual guide map. The resulting binary mask is evaluated against the ground-truth bounding-box mask.

\section{More Qualitative Results Comparison and Analysis}
\label{app:qual}
We provide an interactive qualitative visualization for cross modal scene retrieval and object grounding in our \href{https://scenebind.github.io/}{\texttt{webpage}}. The viewer includes examples across six retrieval directions: Audio to Visual, Visual to Audio, Audio to Text, Text to Audio, Visual to Text, and Text to Visual. For audio queries and retrieved audio samples, we recommend using headphones or speakers that support spatial audio playback.

For each query, the viewer shows the top 3 retrieved samples from SceneBind and compares them with ImageBind~\cite{girdhar2023imagebind}, the strongest zero shot pretrained baseline, and M2D-SigLIP2*, the strongest fine tuned baseline built from M2D-CLAP~\cite{m2d-clap} and SigLIP2~\cite{Tschannen2025SigLIP2M}. Fine tuning details are provided in Appendix Sec.~\ref{app:baseline}. SceneBind results include the global score ($S_{\mathrm{global}}$), the best match object slot score ($S_{\mathrm{obj}}$), and the fused scene score ($S_{\mathrm{scene}}$), where the scores are defined in Appendix Sec.~\ref{app:matching}. These scores make it possible to inspect how global and object level matching contribute to each ranking.

The viewer also shows SceneBind object grounding results on the right. We display all active slots with confidence above 0.05, together with their predicted azimuth, elevation, distance, confidence, and semantic similarity to the best matched ground truth clause. Underlined clauses indicate the best confidence weighted semantic match for each ground truth object clause.

The examples support the analysis in Sec.~\ref{sec:qual}. Across retrieval directions, SceneBind slots often recover foreground objects and events that are important for distinguishing scenes, including both visible objects and sounding sources. SceneBind Matching then compares these object semantics and spatial attributes together with the global scene context. This helps when candidates share similar coarse semantics but differ in object location or composition.

We will include some of the samples in this section and provide case-specific analysis.

\begin{figure}[htb]
\centering
\includegraphics[width=\linewidth]{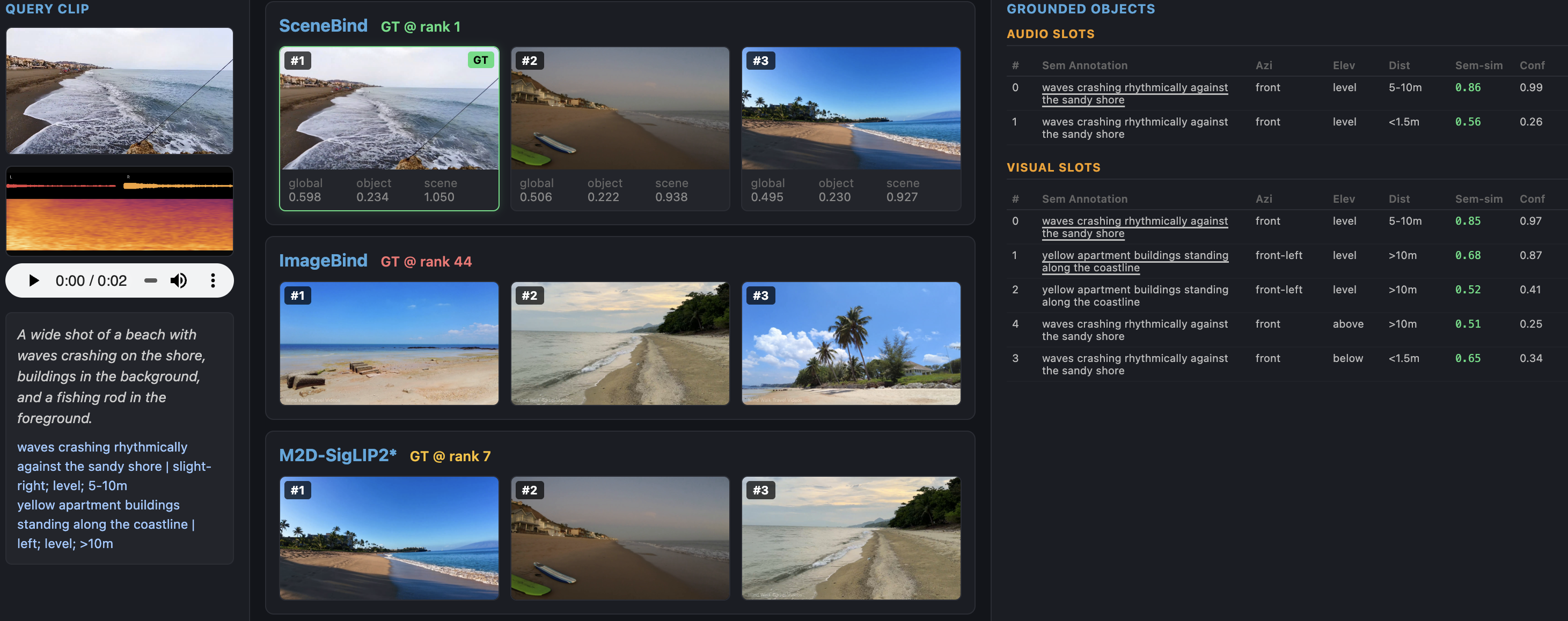}
\caption{\textbf{A$\rightarrow$V beach retrieval.} SceneBind ranks the ground truth first by matching both the global beach semantics and the object level spatial cue of waves, while baseline retrievals capture coarse beach semantics but may miss the spatial layout.}
\label{fig:app-qual-a2v-beach}
\end{figure}

\textbf{Case 1: Spatial Cues Refine Global Retrieval.}
Figure~\ref{fig:app-qual-a2v-beach} shows an A$\rightarrow$V example where the query audio contains waves crashing near the shore. SceneBind grounds an audio slot corresponding to waves in front ($-15$ to $15$ degree azimuth) of the listener, at level elevation and roughly 5-10m distance. The ground truth image is ranked first, with both high global and object scores. The top retrieved SceneBind candidates are all beach scenes with similar layouts, where the shoreline appears in slightly to the front right, consistent with the binaural spatial cues. In contrast, ImageBind and M2D-SigLIP2* retrieve semantically related beach or ocean scenes, but their spatial layouts differ, with the ocean appearing too far to the left or too distant. As a result, the ground truth appears at rank 44 for ImageBind and rank 7 for M2D-SigLIP2*. This example illustrates how SceneBind uses object level spatial cues to refine global semantic retrieval.

\begin{figure}[htb]
\centering
\includegraphics[width=\linewidth]{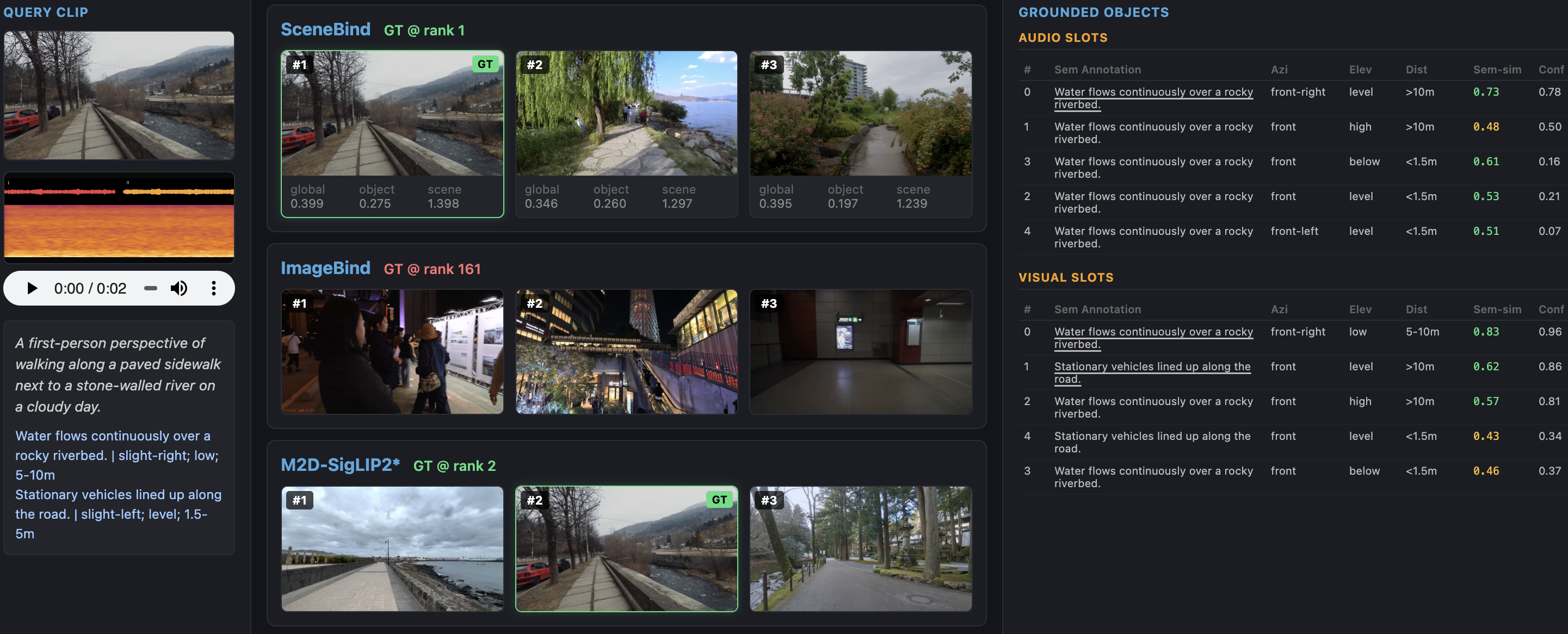}
\caption{\textbf{T$\rightarrow$V river retrieval.} SceneBind matches both the global river road context and object level spatial cues of the riverbed and vehicles, while ImageBind is less sensitive to the detailed text query and M2D-SigLIP2* misses key objects or retrieves weaker layouts.}
\label{fig:app-qual-t2v-river}
\end{figure}

Figure~\ref{fig:app-qual-t2v-river} shows a similar T$\rightarrow$V example. The text query describes a rocky riverbed slightly to the right at 5--10m and vehicles lined up along the road slightly to the left at 1.5--5m. SceneBind grounds the water region on the right and matches the vehicle clause to nearby road objects with high semantic similarity. Its top 3 retrieved samples preserve the same holistic scene context and layout, including the riverbed, road, and surrounding background. In contrast, ImageBind retrieves less relevant scenes for the text query, while M2D-SigLIP2* misses the vehicle in its top result and only retrieves a partially matched riverbed scene at rank 3. This example further shows how object level spatial matching helps distinguish scenes with similar global semantics but different object layouts. 

\begin{figure}[htb]
\centering
\includegraphics[width=\linewidth]{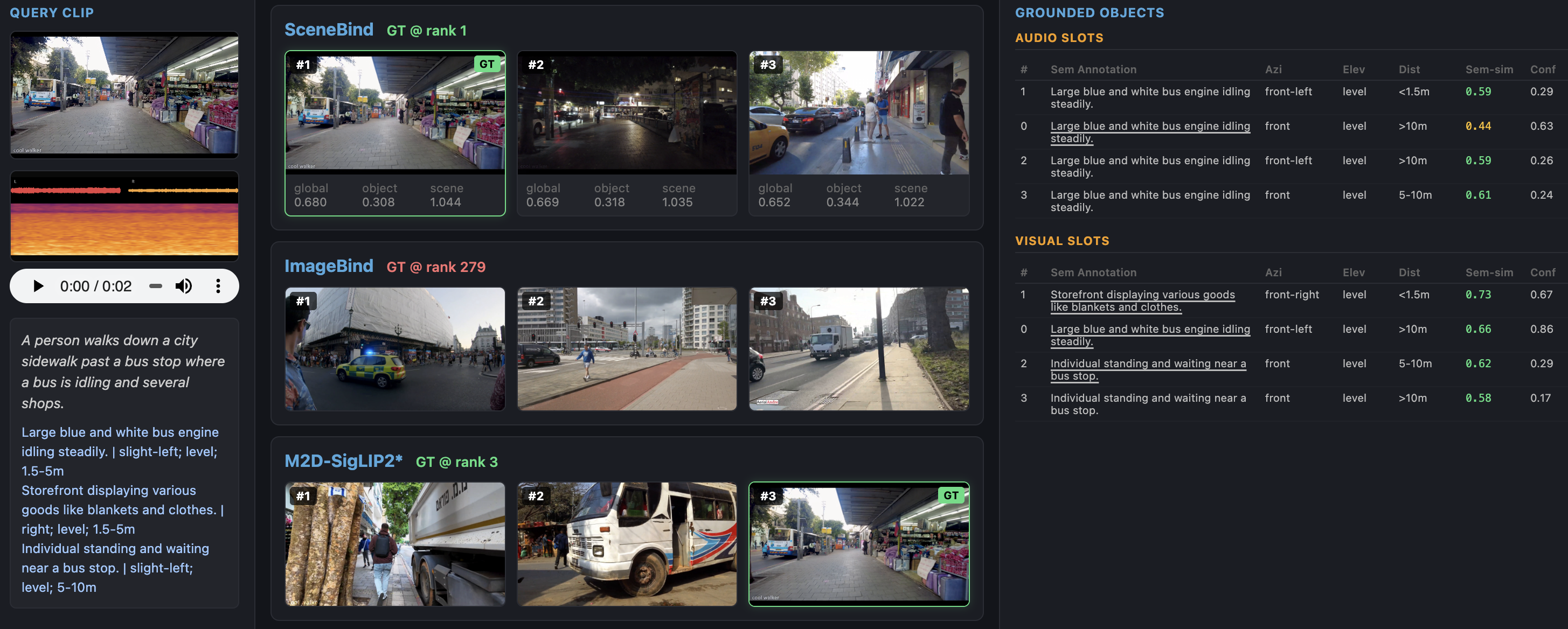}
\caption{\textbf{A$\rightarrow$V moving bus retrieval.} SceneBind uses multiple bus engine relevant slots to represent plausible semantic spatial hypotheses for a moving sound source, retrieving scenes with road and vehicle layouts that better match the binaural audio.}
\label{fig:app-qual-a2v-bus}
\end{figure}

\textbf{Case 2: Multiple Slots Capture Moving Sound Sources.}
Figure~\ref{fig:app-qual-a2v-bus} shows another A$\rightarrow$V example where the query audio contains a bus or car passing from the left toward the front. SceneBind activates multiple bus engine slots with similar semantics but different spatial predictions, such as engine sounds in the front and front-left directions, with medium to far distance and one closer prediction. \textbf{This illustrates the role of object slots: they are not forced to ground a single unique target, but can represent a set of plausible semantic spatial hypotheses, especially for dynamic sources observed over an audio window.} As a result, SceneBind retrieves images with roads and vehicles arranged consistently with the audio layout. In contrast, ImageBind and M2D-SigLIP2* retrieve semantically related street scenes, but with more varied vehicle and lane layouts. This example shows how SceneBind slots expose object-level spatial structure while also revealing uncertainty for moving sources.

\begin{figure}[htb]
\centering
\includegraphics[width=\linewidth]{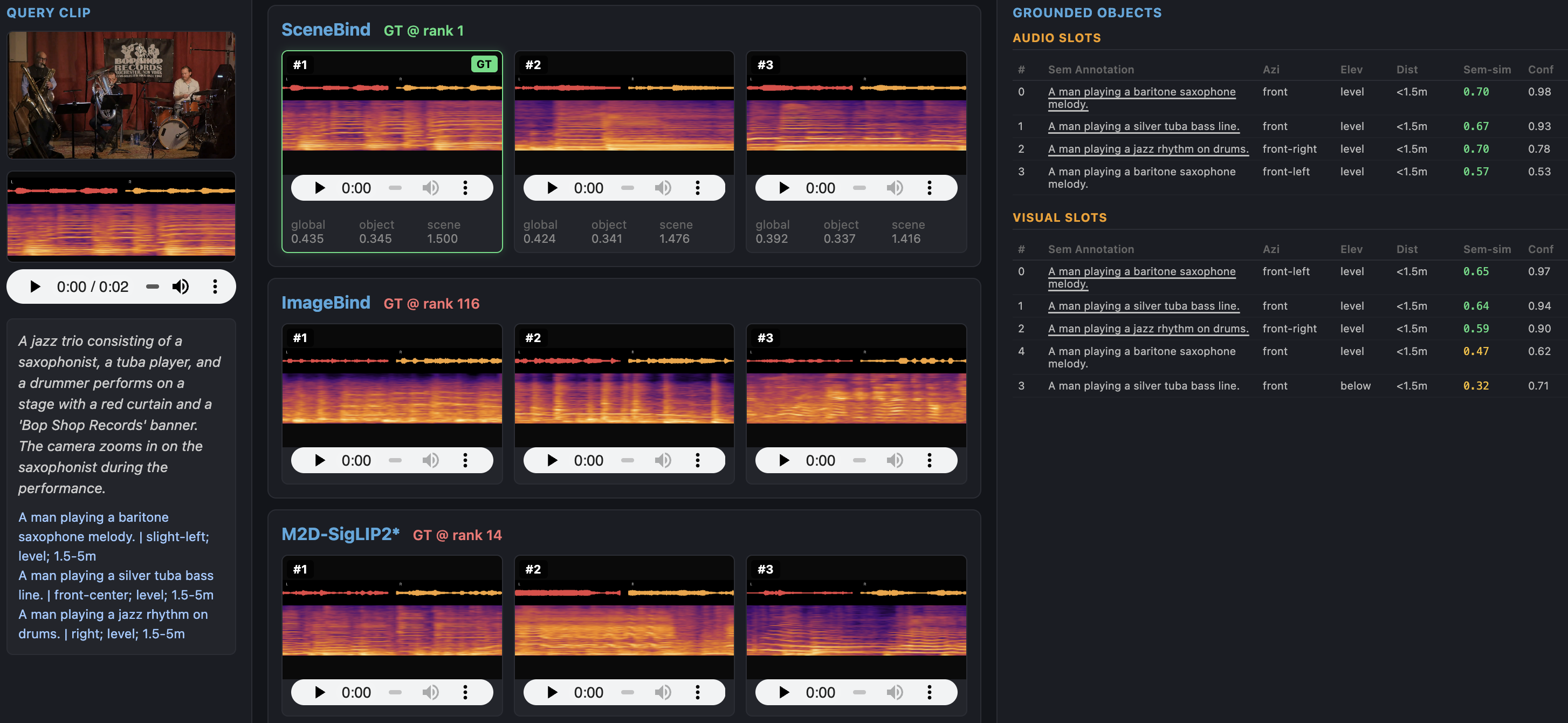}
\caption{\textbf{T$\rightarrow$A jazz retrieval.} SceneBind ranks the ground-truth audio first, and its top retrieved samples show similar jazz performance context, spectrogram and waveform patterns, and object level spatial cues for instruments such as saxophone and drums.}
\label{fig:app-qual-t2a}
\end{figure}

\textbf{Case 3: Audio as the Retrieved Target.}
Figure~\ref{fig:app-qual-t2a} shows a T$\rightarrow$A example from a jazz performance. ImageBind retrieves audio samples with noticeably different spectrograms, although they sound containing drum-like sounds. Their overall scene context and spatial layout differ from the query. According to the samples, ImageBind is not able to adapt well to our text scene domain under the zero-shot setting. After fine tuning, M2D-SigLIP2* better matches the global jazz bar context, but the drum and saxophone locations remain different, which is also reflected in the spectrogram and waveform patterns. In contrast, SceneBind grounds saxophone-related slots in the front and front-left directions, close to the ground-truth saxophone location, and identifies other instruments such as drums and tuba in compatible directions. Though SceneBind ranks the ground-truth audio first, the retrieved audio has spectrogram and waveform patterns closer to the ground truth, and it sounds like describing a similar semantic and spatial layout, such as saxophone in front or front-left and drums slightly to the right.

\begin{figure}[htb]
\centering
\includegraphics[width=\linewidth]{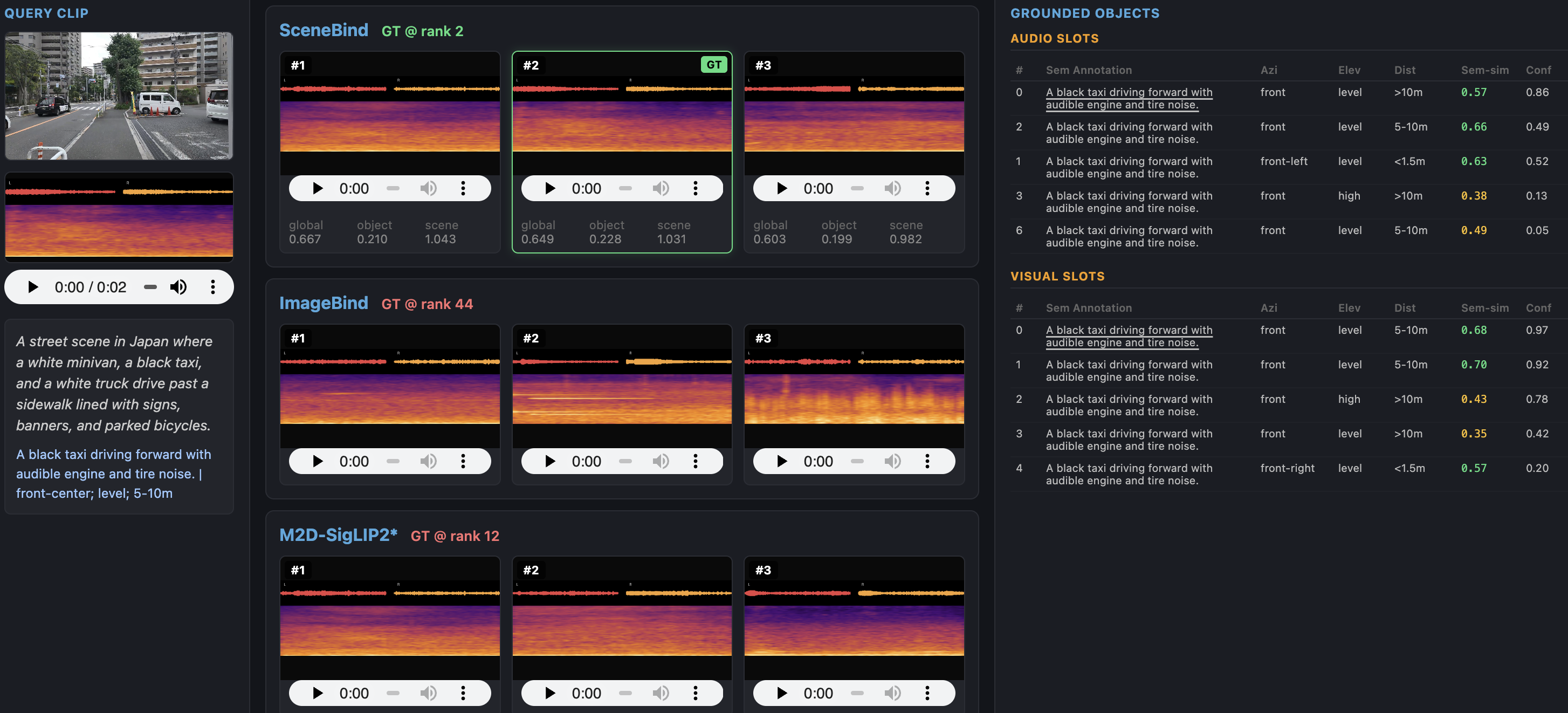}
\caption{\textbf{V$\rightarrow$A street retrieval.} SceneBind retrieves audio with similar street-scene acoustics and frontal vehicle cues. Although the ground truth is ranked second, the top results remain semantically and spatially close, illustrating the challenge of scene retrieval when audio is the target modality.}
\label{fig:app-qual-v2a}
\end{figure}

Figure~\ref{fig:app-qual-v2a} shows a V$\rightarrow$A street example. SceneBind grounds a black taxi in front from the visual slots, while the retrieved audio slots place traffic or vehicle sounds in a similar frontal direction. The top 3 retrieved audios also have similar spectrogram and waveform texture, with street-scene acoustics resembling taxis and traffic flow from the front-left region. Although the ground truth is ranked second, the top-ranked audio remains semantically and spatially close. ImageBind captures the broad street context but shows weaker spatial layout agreement, while M2D-SigLIP2* retrieves traffic sounds with more divergent vehicle directions and sounding object spatial structure.

\textbf{In summary, when audio is the target modality, it can be more challenging because audio provides sparser foreground object evidence than images and thus several candidates in the audio pool may share similar global scene semantics and even spatial layout.}

\section{Limitations and Future Works}
\label{app:limitations}
\textbf{Data Scale and Annotation Quality.}
SceneBind relies on curated spatially aligned audio--visual--text data with structured semantic and spatial annotations. While our binaural dataset provides a strong starting point, its scale and diversity remain limited compared to web-scale semantic datasets. Real-world scenes exhibit long-tail event distributions, complex acoustic conditions, and ambiguous spatial cues that are difficult to capture with current annotation pipelines. In addition, our automatic annotation step is guided by Gemini, which does not directly perceive binaural spatial cues. Its spatial estimates are inferred mainly from visual evidence, which can bias annotations toward visible or front-facing objects and may miss or hallucinate off-screen or rear-field events. Although we mitigate these issues with cross-modal verification, data balancing, and human review for evaluation, improving automatic spatial annotation quality remains an important direction. Future work will scale data collection with richer in-the-wild sources, strengthen multimodal filtering and human-in-the-loop refinement, and explore semi-supervised or self-supervised learning to reduce reliance on explicit spatial labels.

\textbf{Temporal Modeling.}
The current formulation operates on short temporal windows (e.g., a single visual frame and $\sim$2s audio), which limits the ability to capture temporal dynamics such as object motion, evolving interactions, and long-range semantic-spatial representation. Extending SceneBind to incorporate temporal integration over longer and flexible time horizons is an important direction. This includes modeling continuous view or object trajectories, temporally consistent slot representations, and cross-time semantic--spatial reasoning for dynamic scenes.

\section{Broader Impacts}
\label{app:broader_impacts}
SceneBind studies in-the-wild semantic-spatial omni-modal representation learning, aiming to move multimodal perception beyond recognizing what is present toward understanding where objects and events are located in 3D space. Such omnidirectional spatial perception can support more natural spatial reasoning in assistive systems, wearable devices, AR/VR interaction, multi-modal navigation, human-robotic interaction, and embodied agents. The representation can also serve as a structured interface for larger systems: combined with language models, it may support natural spatial question answering and instruction following; combined with generative or world models, it may help construct semantically and spatially aligned scene representations, or provide a reward or critic signal for evaluating multimodal spatial consistency and improving omni-modal generative models.

At the same time, models that infer spatial structure from audio-visual input can create risks if deployed without safeguards. Potential misuse includes surveillance, unauthorized tracking, or context inference about people and environments without consent. SceneBind also inherits limitations and biases from its pretrained encoders and training data, and its spatial predictions may be incorrect in ambiguous or safety-critical settings. To reduce these risks, deployments should communicate uncertainty and failure modes, follow privacy-preserving data collection and evaluation practices, and restrict use in sensitive applications involving biometric data, private spaces, or real-time monitoring. Future work should further study robustness, interpretability, and consent-aware evaluation protocols for spatial omni-modal perception and reasoning.


\end{document}